\documentclass{article}

\usepackage[english]{babel}

\usepackage[letterpaper,top=2cm,bottom=2cm,left=3cm,right=3cm,marginparwidth=1.75cm]{geometry}

\usepackage{amsmath}
\usepackage{graphicx}
\usepackage[colorlinks=true, allcolors=blue]{hyperref}
\usepackage{makecell}
\usepackage{bbding}
\usepackage{hyperref}

\title{Lightweight Full-Convolutional Siamese Tracker}
\author{Yunfeng Li, Bo Wang, Xueyi Wu, Zhuoyan Liu, Ye Li\\
        Harbin Engineering University \\
}
\begin{document}
\maketitle

\begin{abstract}
Although single object trackers have achieved advanced performance, their large-scale models hinder their application on limited resources platforms. Moreover, existing lightweight trackers only achieve a balance between 2-3 points in terms of parameters, performance, Flops and FPS. To achieve the optimal balance among these points, this paper proposes a lightweight full-convolutional Siamese tracker called LightFC. LightFC employs a novel efficient cross-correlation module (ECM) and a novel efficient rep-center head (ERH) to improve the feature representation of the convolutional tracking pipeline. The ECM uses an attention-like module design, which conducts spatial and channel linear fusion of fused features and enhances the nonlinearity of the fused features. Additionally, it refers to successful factors of current lightweight trackers and introduces skip-connections and reuse of search area features. The ERH reparameterizes the feature dimensional stage in the standard center-head and introduces channel attention to optimize the bottleneck of key feature flows. Comprehensive experiments show that LightFC achieves the optimal balance between performance, parameters, Flops and FPS. The precision score of LightFC outperforms MixFormerV2-S on LaSOT and TNL2K by 3.7 \% and 6.5 \%, respectively, while using 5x fewer parameters and 4.6x fewer Flops. Besides, LightFC runs 2x faster than MixFormerV2-S on CPUs. In addition, a higher-performance version named LightFC-vit is proposed by replacing a more powerful backbone network. The code and raw results can be found at \href{https://github.com/LiYunfengLYF/LightFC}{https://github.com/LiYunfengLYF/LightFC}.
\end{abstract}

\section{Introduction}
Single object tracking (SOT) is a fundamental task in computer vision, which aims to obtain the track of a target by utilizing the target’s appearance template in a sequence of images or a video. In recent years, while single object trackers have made surprising performance improvements, these trackers are designed on large network structures with many parameters and Flops. Deploying them on devices with limited resources is challenging because of high cost.

A lightweight tracker needs to achieve an optimal balance between performance, parameters, Flops and FPS. Full-convolutional lightweight trackers with fewer parameters and Flops, such as LightTrack  \cite{LightTrack} and FEAR \cite{FEAR}, perform insufficiently. For example, the AUC of LightTrack \cite{LightTrack} and FEAR \cite{FEAR} on LaSOT \cite{LaSOT} is 7 \% and 6.8 \% lower than that of MixFormerV2-S \cite{MixFormerV2}, respectively. To improve the performance of the lightweight tracker, efficient attention mechanisms are introduced to design the trackers. E.T.Track \cite{E.T.Track} and MixFormerV2-S \cite{MixFormerV2} achieve significant performance improvements. However, the use of the attention mechanism results in a significant increase in both params and Flops. For instance, E.T.Track \cite{E.T.Track} and MixFormerV2-S \cite{MixFormerV2} is 3.5x and 8x more params, 3x and 8x more Flops than LightTrack \cite{LightTrack}, respectively. As a result, the current lightweight trackers only achieve a balance of two or three points rather than the optimal four points.

This paper aims to design an efficient tracker that effectively balances these four points. To minimize the number of parameters and Flops, a full-convolutional tracking pipeline is selected as the baseline. To achieve better performance, this paper focuses on improving the model’s feature expressiveness. An effective network design serves as the foundation for other improvement strategies, including training methods. Thus, the critical problem designs an efficient feature fusion module and an efficient prediction head to boost the tracker’s performance.

\begin{figure}
	\centering
        \includegraphics[width=10cm]{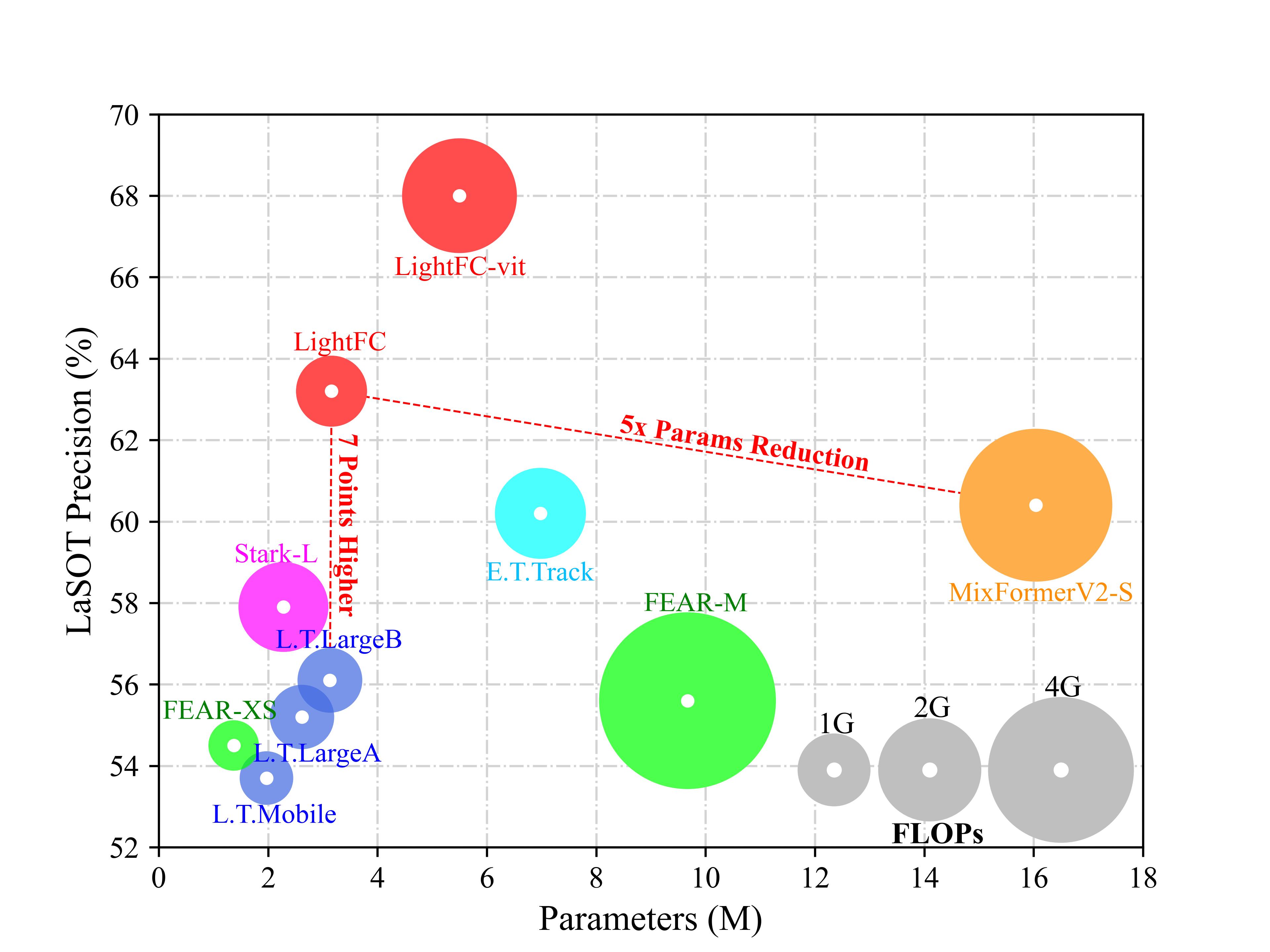}
	\caption{Comparisons with state-of-the-art lightweight trackers in terms of precision performance, parameters and model Flops on LaSOT \cite{LaSOT} benchmark. The circle diameter is in proportion to the number of model Flops. L.T.Track represents LightTrack \cite{LightTrack}, Stark-L represents Stark-Lightning \cite{Stark}. The proposed LightFC and LightFC-vit are superior to LightTrack \cite{LightTrack}, Stark-Lightning \cite{Stark}, FEAR \cite{FEAR}, E.T.Track \cite{E.T.Track} and MixFormerV2-S \cite{MixFormerV2}. Best viewed in color.}
	\label{fig:fig1}
\end{figure}

Therefore, a lightweight full-convolutional tracking pipeline LightFC is proposed. LightFC comprises two novel components, an Efficient Cross-Correlation Module (ECM) and an Efficient Rep-Center-Head (ERH). The ECM fuses the space and channels of fused features and then enhances the nonlinearity of the fused feature. The successful factors of the feature fusion modules from classic lightweight trackers are also integrated to further improve the ECM. The ERH introduces reparameterization and channel attention to optimize the feature representation bottleneck between the first and second convolutional blocks of each branch of the center-head \cite{centerhead}. These two modules significantly improve the model’s feature representation with minimal expense (LightFC: 3.2 M parameters, while LightTrack-LargeB \cite{LightTrack}: 3.1 M parameters). Higher performance is attained by another version called LightFC-vit, which substitutes the more powerful TinyViT \cite{tinyvit} for the MobileNetv2 \cite{MobilNetV2} backbone network. As shown in Fig. 1, LightFC achieves better performance than conv-based lightweight trackers, even outperforming the most advanced attention-based lightweight trackers. Besides, LightFC achieves fewer parameters and Flops than attention-based lightweight trackers. LightFC outperforms MixFormerV2-S \cite{MixFormerV2} by 3.7 \% and 6.5 \% in precision and 1.2 \% and 2.6 \% in AUC scores on TrackingNet \cite{TrackingNet} and TNL2K \cite{TNL2K}, respectively, while using 5x fewer parameters and 4.6x fewer Flops. In addition, LightFC is 2x faster than MixFormerV2-S \cite{MixFormerV2} on CPUs. LightFC-vit outperforms LightFC in terms of precision and AUC score on LaSOT by 4.8 \% and 3 \%, respectively. 

The paper’s main contributions are:

1. An Efficient Cross-Correlation Module (ECM) is designed based on the characteristics of pixel-wise correlation relationship modeling. Subsequently, a spatial and channel fusion (SCF) unit and an inverted activation block (IAB) are added to combine the local space and channel features and enhance the model’s nonlinearity, respectively. In addition, the success factors of existing feature fusion modules are analyzed and applied to further improve the performance of ECM.

2. An Efficient Rep-Center-Head (ERH) is proposed to eliminate the bottleneck of feature flow representation for center-head \cite{centerhead} by utilizing the reparameterization technique. To improve feature representation in its space and channels, the first convolutional block is reparameterized on each branch of the center-head \cite{centerhead}. In addition, an SE \cite{SE} module is added between the first and second blocks of to further optimize the bottleneck of key feature expression.

3. This paper proposes two Lightweight full-convolutional siamese trackers, namely LightFC and LightFC-vit. Comprehensive experiments show that they achieve the state-of-the-art performance on several benchmarks, while maintaining a good balance between parameters, Flops, FPS, and performance.

\section{Related Work}

\subsection{Lightweight Convolutional Networks}
Lightweight convolutional neural networks (CNNs) have made significant progress. Many efficient model designs have been proposed to improve the performance of lightweight CNNs. Separable convolution proposed by MobileNetV1 \cite{MobilNetV1} decomposes standard convolution into depth-wise convolution and point-wise convolution to reduce size and latency. Channel shuffle proposed by ShuffleNets \cite{ShuffleNetV1}\cite{ShuffleNetV2} reduces cost and achieves better accuracy. Inverted residual block proposed by MobileNetV2 \cite{MobilNetV2} is a classic and efficient module design in lightweight networks. Besides, it also plays an important role in modern large-scale network design, such as conv-based FFN \cite{CMT} in transformer networks. Many efficient networks, such as MobileNetV3 \cite{MobilNetV3} and EfficientNets \cite{EfficientNet}, use network architecture search (NAS) to design their model. Structure reparameterization which is employed by RepVGG \cite{RepVGG}, MobileOne \cite{MobileOne} and RepViT \cite{RepViT} provides a technique for improving model performance without adding additional parameters. It uses additional convolutional kernels during training to enhance the feature expression of the model and fuses multiple convolutional kernels into a single convolutional kernel during inference. Due to its effectiveness, reparameterization has been introduced in object detection such as YOLOv6 \cite{YOLOv6} and YOLOv7 \cite{YOLOv7}, but it has not yet received much attention in SOT.

\subsection{Siamese Trackers}
Siamese network has become the basic architecture in trackers due to its simplicity and efficiency. Powerful feature fusion networks and prediction heads are essential components of Siamese-based trackers. In its pioneering work, SiamFC \cite{SiamFC} employed naïve correlation as a fusion operation and prediction head. SiamRPN \cite{SiamRPN} and SiamRPN++ \cite{SiamRPNpp} improve SiamFC by employing the RPN network as the prediction head. SiamAttn \cite{SiamAttn} introduced deformable convolutional attention to improve feature fusion and proposed a region refinement module to obtain a more accurate box. As anchor-free trackers, SiamBAN \cite{SiamBAN} and SiamCAR \cite{SiamCAR} employed depth-wise correlation to fuse the features of the template and search area. Then they used different types of fully convolutional network (FCN) or multi-layer perceptron (MLP) as their prediction head. SiamOAN \cite{SiamOAN} proposed a conv-based channel attention module to enhance the features of the template and search area, then it used naïve depth-wise correlation for further features fusion. To model long-term context, dynamic context, and preliminarily model the relationship between template and search area, SiamAGN \cite{SiamAGN} proposed convolutive self-attention-like and cross-attention-like module. Subsequently, depth-wise correlation is used for further fusion. SiamBAN-ACM \cite{SiamBAN_ACM} proposed an asymmetric convolution in which the convolution operation is decomposed into two mathematically equivalent operations to improve the depth-wise correlation. 

In contrast to them, LightFC focuses on enhancing the feature representations after fusion, rather than before fusion or during fusion. In addition, LightFC performs a detailed analysis and optimization of the center-head’s feature representation bottleneck  \cite{centerhead} rather than simply applying an FCN as the prediction head.

With the introduction of the transformer architecture in SOT, trackers based on attention mechanisms have achieved better feature fusion, including self-attention \cite{Stark}, self-attention and cross-attention \cite{TransT}, improved attention \cite{AiaTrack}, visual transformer \cite{OSTrack}, mixed-attention \cite{MixFormer} have utilized attention mechanisms to achieve more powerful feature fusion. 

Although these transformer-based models are difficult to directly apply to lightweight trackers, tricks in these attention models as nonlinear layers and skip-connection can provide reference for designing full-convolutional fusion modules.

\subsection{Lightweight Trackers}
Alpha-Refine \cite{Alpha_Refine} discussed the performances of naïve correlation, depth-wise correlation, and pixel-wise correlation in detail in a small-scale tracker, and demonstrated the effectiveness of pixel-wise correlation. LightTrack \cite{LightTrack} employed Neural Architecture Search (NAS) to search a convolutional architecture that is both lightweight and efficient. Its feature fusion module only contains pixel-wise correlation and an SE \cite{SE} module. Its prediction head is an FCN which consists of two convolutional branches. FEAR \cite{FEAR} used separable convolutional blocks to enhance the feature expressiveness of fused features and proposed a more efficient hand-crafted convolutional structure and a template update approach. It used two branches with the same structure for classification and regression, respectively.
Stark-Lightning \cite{Stark} adopted Rep-VGG \cite{RepVGG} as the backbone for feature extraction and a single transformer encoder is used for feature fusion. E.T.Track \cite{E.T.Track} proposed exemplar attention to improve the predicted head of LightTrack \cite{LightTrack}. MixFormerV2-S \cite{MixFormerV2} proposed a single-stream full-transformer tracking architecture. These attention-architectured lightweight trackers perform competitively, but a major shortcoming is that they remain expensive and heavy.
For the features fusion module, LightFC employs the LightTrack \cite{LightTrack} and E.T.Track’s \cite{E.T.Track} fusion module as the baseline. Compared to Alpha-Refine \cite{Alpha_Refine} and LightTrack \cite{LightTrack}, LightFC improves feature representation and enhances model nonlinearity. In addition, the reuse of search area features also supplements the lost target semantic information. Compared to FEAR \cite{FEAR} , LightFC only uses one branch consisting of the fusion module and the predict head, rather than using two branches with the same structure. LightFC improves the feature representation of the prediction head at no additional cost as compared to other methods such as Alpha-Refine \cite{Alpha_Refine}, LightTrack \cite{LightTrack}, FEAR \cite{FEAR}, Stark-Lightning \cite{Stark} and MixFormerV2-S \cite{MixFormerV2}. Compared to E.T.Track \cite{E.T.Track}, the fully convolutional prediction head of LightFC has fewer parameters and flops.

\section{LightFC}
\begin{figure}
	\centering
        \includegraphics[width=16cm]{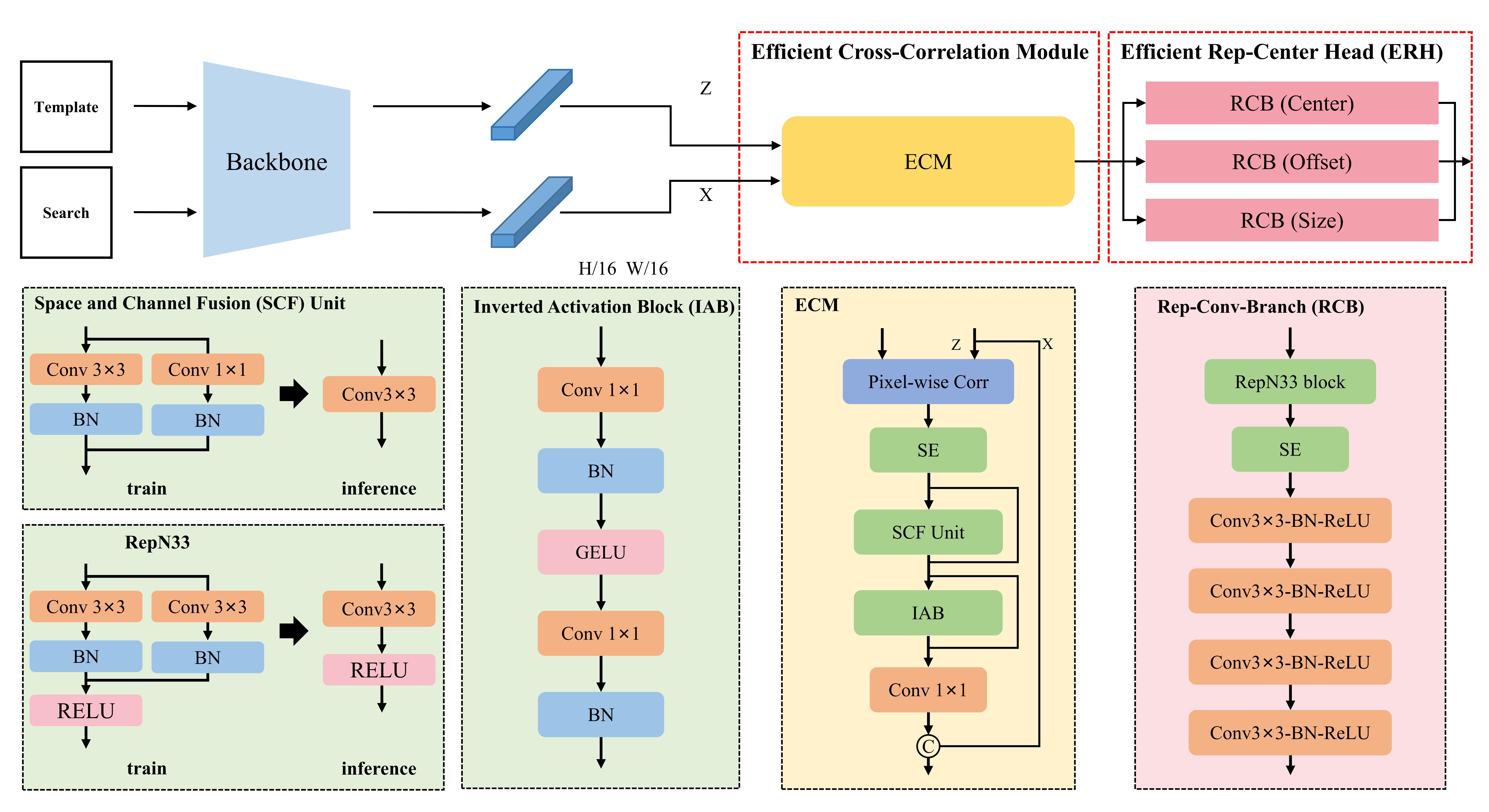}
	\caption{The overview of LightFC architecture. H and W represent the width and height of the input image. Z and X denote the template and search area features, respectively.}
	\label{fig:fig2}
\end{figure}

The key factor of designing LightFC is to improve the feature representation of the full-convolutional model. Fig.\ref{fig:fig2} shows the overall architecture of LightFC, which consists of a backbone, an Efficient Cross-Correlation Module (ECM), and an Efficient Rep-Center Head (ERH). This work can be summarized as four parts. First, the most suitable backbone network for LightFC’s tracking pipeline is identified from a broad range of efficient networks. Second, this paper analyzes the success factors of current feature fusion modules for lightweight trackers and proposes three hypotheses: adding nonlinear blocks, skip-connections, and reusing search area features to develop the ECM. Third, this paper analyzes the bottleneck of feature flow representations in the center-head \cite{centerhead} and introduces reparameterization and channel attention to propose the ERH. Fourth, out of a large number of intersection-over-union (IoU) loss functions, the paper selects the most suitable box IoU loss function. A detailed analysis of the effectiveness of each part is provided in Section \ref{sec:subsection4.3}.

\subsection{Backbone}
A lightweight and efficient backbone is essential for improving the performance of a lightweight tracking pipeline. \cite{LightTrack} and \cite{OSTrack} demonstrate the effectiveness of using a pretrained backbone to improve the tracker’s performance. Therefore, this paper conducts a comprehensive performance analysis of current pretrained lightweight backbones in  LightFC’s tracking pipeline  to select the most suitable one. Refer to Section \ref{sec:subsection4.3.1}.Mobilentv2 \cite{MobilNetV2} and TinyViT \cite{tinyvit} are selected as baselines because they exhibit the best performance in convolutional and transformer backbone networks, respectively. The input of backbone is a pair of images, namely, the template image $z\in R^{3\times H_{z}\times W_{z}}$ and the search area images $x\in R^{3\times H_{x}\times W_{x}}$. Then they are fed into the backbone to extract features $z_{f}\in R^{3\times H_{zf}\times W_{zf}}$ and $x_{f}\in R^{3\times H_{xf}\times W_{xf}}$, where $H_{if}, W_{if}=H_{i}/S, W_{i}/S, i\in (z,x)$, $C=96$ is output channel, $S=16$ is stride.

\subsection{Efficient Cross-Correlation Module} \label{sec:subsection3.2}
\begin{figure}
	\centering
        \includegraphics[width=16cm]{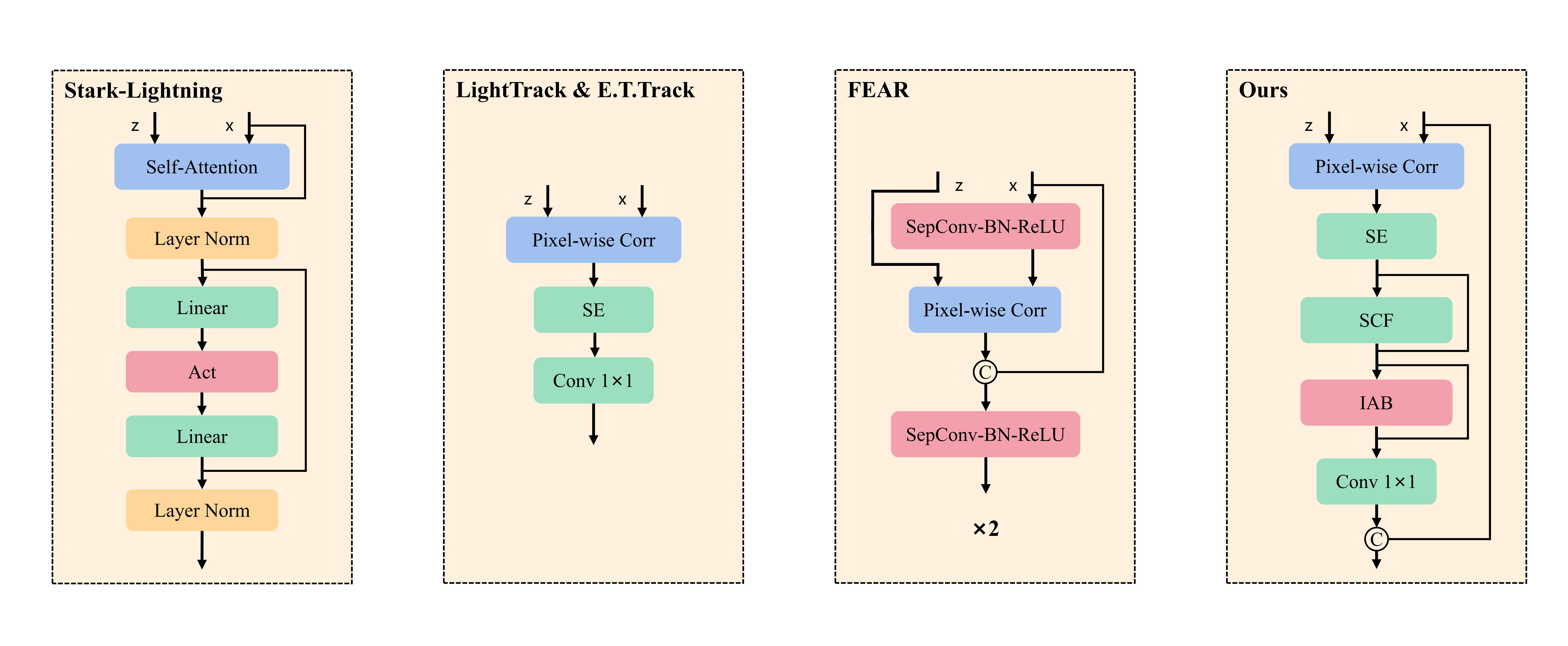}
	\caption{Comparison of feature fusion modules for current lightweight trackers.}
	\label{fig:fig3}
\end{figure}

First, the design of feature fusion modules for Stark-Lightning  \cite{Stark}], LightTrack \cite{LightTrack}, E.T.Track \cite{E.T.Track} and FEAR \cite{FEAR} is reviewed. Nonlinear blocks like the Linear-Activation-Linear block and the SepConv-BN-ReLU block are crucial in enhancing the nonlinearity of the model, as shown in Fig.\ref{fig:fig3} in Stark-Lightning \cite{Stark} and FEAR \cite{FEAR}. Besides, the skip-connections Stark-Lightning \cite{Stark} preserves original information. Additionally, both Stark-Lightning \cite{Stark} and FEAR \cite{FEAR} reuse the search area feature to supplement target semantic information. Even though features fusion module used by LightTrack \cite{LightTrack} and E.T.Track \cite{E.T.Track} is very efficient, it does not further consider improving the nonlinearity of features and optimizing the use of existing information to improve feature representation. 

Therefore, this work employs pixel-wise correlation and an SE [53] module as a baseline and makes the following assumptions to improve the feature representation:
1. The addition of nonlinear blocks could enhance the nonlinearity of the feature fusion module.
2. The introduction of skip-connections could preserve the semantic information of original features \cite{resnet}.
3. The reuse of search area features could supplement the target’s appearance semantic information that is lost during feature fusion \cite{desnet}.
The feature fusion module baseline is described as:

\begin{equation}
X_{f(z,x)}=SE(\{x_{f(z,x)}^j|x_{f(z,x)}^j=z^{j}_{f}*x_{f}\}_{j\in \{1,...,H_{zf}\times W_{zf}\}}) \in R^{C_{f}\times H_{xf}\times W_{xf}} \label{XX}
\end{equation}
where $C_{f}=H_{zf}\times W_{zf}$, $SE$ denotes SE function and $*$ denotes naïve correlation.

\begin{figure}
	\centering
        \includegraphics[width=8cm]{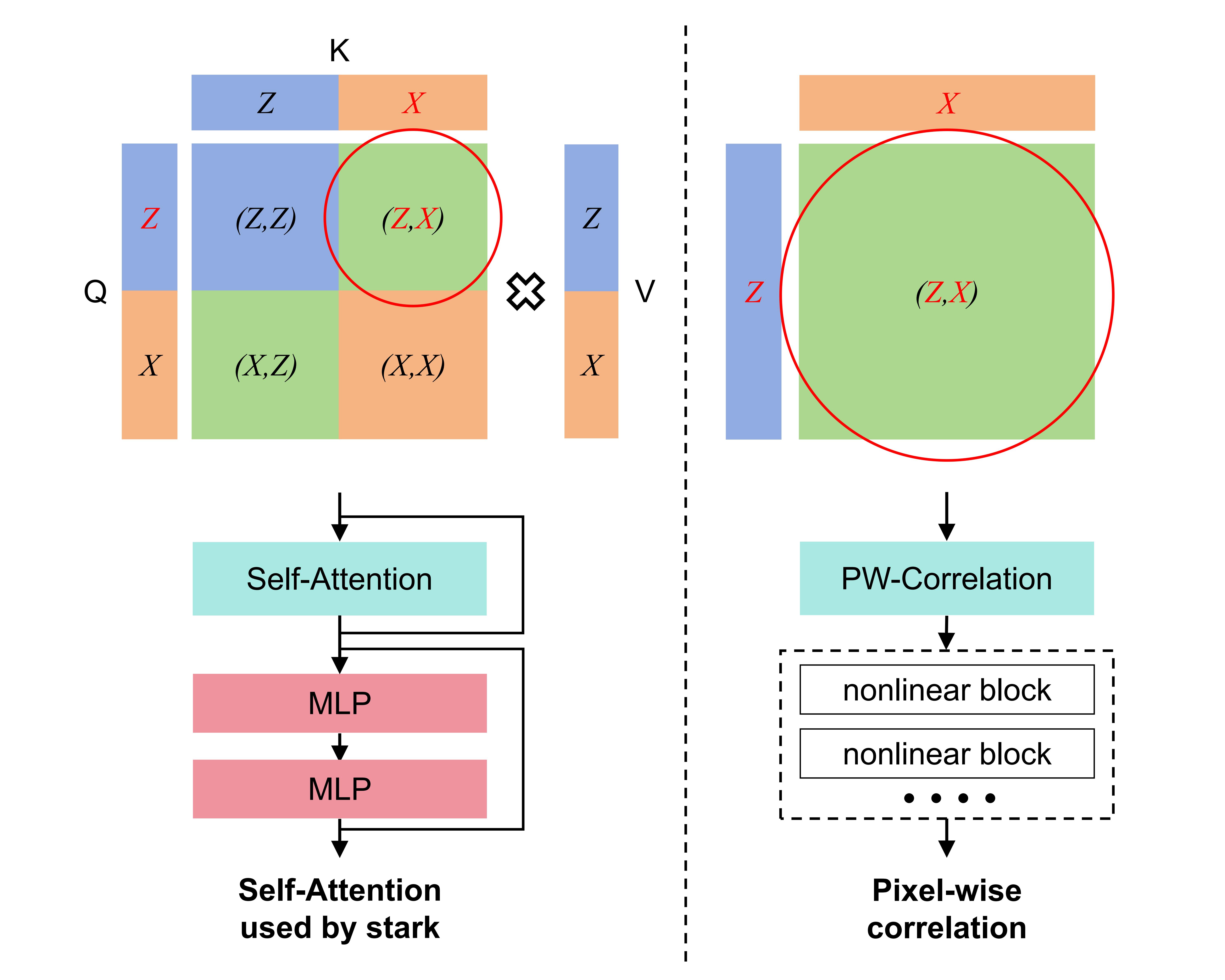}
    	\caption{llustration of how we design the ECM architecture based on the self-attention module used by the trackers.}
	\label{fig:fig4}
\end{figure}

Pixel-wise correlation treats each point in the template feature as a convolution kernel and applies convolution to the entire search area features. Thus, each point in the template features interacts with the global search area features. Pixel-wise correlation operates similarly to a conv-based cross-attention, similar to the tracker’s self-attention by modeling the relationship between the template and the search area, as shown in Fig.\ref{fig:fig4}.Therefore, following the first factor above and the architecture design of the attention-like module, two nonlinear blocks are introduced behind the pixel-wise correlation layer to improve the model’s performance.

Each layer of the fused feature corresponds to the naïve correlation map between each point on the template and the entire search region. The local spatial information of each layer’s response peak lacks further refinement, and the fused features operate independently, resulting in a leak of information exchange between the layers. To address the issues, this paper first introduces the space and channel fusion (SCF) unit to linearly fuse local spatial information and channel information in the fused features. The SCF consists of a 3x3 Conv-BN branch $E_{33}$ and a 1x1 Conv-BN branch$E_{11}$.$E_{33}$ maps $x_{f(z,x)}$ from $R\in R^{C_{f}\times H_{xf}\times W_{xf}}$ to $R\in R^{C_{f}\times H_{xf}\times W_{xf}}$ and models the local spatial information. $E_{11}$ makes the same dimensional transformation as $E_{33}$ and fuses the channel information. The structure of SCF can be further equivalent to a convolutional kernel by structural reparameterization technique. Moreover, skip-connection is introduced to remain the original fusion feature. The process of SCF can be written as follows:

\begin{equation}
\Tilde{E}_{33}=\phi (E_{33}+E_{11})\label{XX}
\end{equation}

\begin{equation}
x_{scf}=E_{33}(x_{f(z,x)}+E_{11}(x_{f(z,x)})+x_{f(z,x)}=\Tilde{E}_{33}(x_{f(z,x)})+x_{f(Z,x)}\label{XX}
\end{equation}
where $\phi$ denotes reparameterization transformation.

To enhance the model’s nonlinearity, an inverted activation block (IAB) is introduced. IAB contains two different 1x1 Conv-BN branches, namely $E_{11}^{up}$ and $E_{11}^{down}$, and a nonlinear Gelu activation function. $E_{11}^{up}$ is used to map $x_{scf}$ from $R^{C_{f}\times H_{xf}\times W_{xf}}$ to $R^{(\alpha \times C_{f})\times H_{xf}\times W_{xf}}$, where $\alpha=2$ is channel expansion rate. And $E_{11}^{down}$ is used to reduce dimension from $\alpha \times C_{f})$ to $C_{f}$. Skip-connection is also introduced. The detailed process is described as follows:

\begin{equation}
x_{iab}=E_{11}^{down}(Gelu(E_{11}^{up}(x_{scf})))+x_{scf}\label{XX}
\end{equation}

Following the design of LightTrack \cite{LightTrack} and E.T.Track \cite{E.T.Track}, a 1x1 Conv branch $E_{11}^{adj}$ is used to increase dimension from $C_{f}$ to $C_{fusion}$, where $C_{fusion}=96$. Besides, feature concatenation is introduced to reuse the search area features for supplementing semantic information that is lost during feature fusion.

\begin{equation}
x_{fusion}=E_{11}^{adj}(x_{iab})\textcircled{c} x_{f}\label{XX}
\end{equation}
where $x_{fusion}$ denotes output features of the ECM and $\textcircled{c}$ denotes tensor concatenate.

\subsection{Efficient Rep-Center Head}
There are two classic types of classification heads. One approach involves outputting a box end-to-end, as seen in the corner-head utilized by the Stark family \cite{Stark}. Another way involves outputting boxes with confidence scores and selecting the one with the highest confidence score as the output. For instance, the center-head \cite{centerhead} used by OSTrack \cite{OSTrack} outputs a response map, an offset map, and a size map before decoding the output box. In contrast, the MLP head used by TransT \cite{TransT} outputs 1024 boxes directly. NeighborTrack \cite{NeighborTrack} and UOSTrack \cite{UOSTrack} demonstrate that tracker performance can be enhanced by reusing candidate boxes. This paper selects the FCN-based Center-Head based as prediction head baseline in order to utilize other well-designed post-processing methods to additionally improve the performance of LightFC. 

Standard center-head \cite{centerhead} has three branches with the same structure. Each branch contains five Convolutional, Batch-Normalization, ReLU (Conv-BN-ReLU or CBR) blocks. Each CBR of a branch can be defined as $F_{33}^{i},i\in \{1,2,3,4,5\}$. Different from $E_{33}$, it has a nonlinear activate function. $F_{33}^{1}$ increases feature dimension from $C_{fusion}$ to $C_{head}=256$, where $C_{head}$ is a hyper-parameter as same in Stark \cite{Stark}. $F_{33}^{i},i\in \{2,3,4,5\}$ reduce feature channels by half. The last block of each branch outputs a response map. Center-head \cite{centerhead} has a feature representation bottleneck between $F_{33}^{1}$ and $F_{33}^{i},i\in \{2,3,4,5\}$ which limits the transmission of key information in the network.

To optimize the feature flow bottleneck in the center-head \cite{centerhead} the impact of various reparameterization kernels on the center-head \cite{centerhead} is investigated in detail, inspired by RepVGG \cite{RepVGG} and proposes an Efficient Rep-Center Head. In particular, a Rep-Conv-Branch (RCB) is designed to replace the standard branch of the center-head. Specifically, a RepN33 block [Illustrated in Fig.\ref{fig:fig8}] $\Tilde{F}_{33}^{1}$ is introduced to replace $F_{33}^{1}$. During training, $\Tilde{F}_{33}^{1}$ is composed of two $F_{33}^{1}$ with different parameters. During testing, they are merged into a block with the same structure as $F_{33}^{1}$. In addition, an SE \cite{SE} module is added between $\Tilde{F}_{33}^{1}$ and $F_{33}^{2}$ to further enhance the expression of key information. Using reparameterization techniques on $F_{33}^{i},i\in \{2,3,4\}$ reduce model performance. Therefore,$F_{33}^{i},i\in \{2,3,4\}$ is preserved. Compared to the conventional branch of center-head \cite{centerhead}, RCB enhances the bottleneck of essential feature expression. The function of RCB can be written as follows:

\begin{equation}
\Tilde{F}_{33}^{1}=\phi(F_{33}^{1}+\bar{F}_{33}^{1})\label{XX}
\end{equation}

\begin{equation}
output_{RCB}=F_{33}^{5}(F_{33}^{4}(F_{33}^{3}(F_{33}^{2}(SE(\Tilde{F}_{33}^{1}(x_{fusion}))))))\label{XX}
\end{equation}
where $\phi$ denotes reparameterization transformation.

The weight focal loss \cite{weightfocalloss} is used to train the classification branch. And the l1 loss and Wise-IoU loss \cite{wiou} are used for box regression. The total training loss of LightFC is:

\begin{equation}
    L_{total}=L_{cls}+\lambda_{iou}L_{iou}+\lambda_{l_{1}}L_{1}
\end{equation}
where $\lambda_{iou}=2$ and $\lambda_{l_{1}}=5$ are hyper-parameters in this paper as same in \cite{Stark}\cite{OSTrack}.

\section{Experiments}

\subsection{Implementation Details}
LightFC is developed based on Ubuntu 20.04, using Python 3.9, and Pytorch 1.13.0. The tracker is trained with 2 NVIDIA RTX A6000 GPUs over 400 epochs, which took roughly 57 hours. The training set includes LaSOT \cite{LaSOT}, TrackingNet \cite{TrackingNet}, GOT10k \cite{GOT10k} and COCO \cite{COCO}. For each epoch, it samples 60000 image pairs to train the model, with a total batch-size of 64. LightFC uses the AdamW optimizer with a total learning rate of 0.0001 for the backbone and 0.001 for other parameters and a weight decay of 0.0001. The total learning rate decreases to 0.0001 after the 160th epoch. Following the training paradigm of \cite{Stark}\cite{OSTrack}, it only employs Brightness Jitter, Grayscale and Random Horizontal Flip data augmentation methods. The training settings of LightFC-vit are the same as LightFC.

This paper follows the one-pass evaluation (OPE) protocol proposed by OTB100 [37] to test our tracker on the benchmarks LaSOT \cite{LaSOT}, TrackingNet \cite{TrackingNet}, TNL2K \cite{TNL2K}, OTB100 \cite{OTB100}, UAV123 \cite{UAV123}, TC128 \cite{TC128}, UOT100 \cite{UOT100} and UTB180 \cite{UTB180}. And success (AUC), precision (P) and norm-precision (PNORM) are employed to evaluate the performance of the trackers on these benchmarks. On the VOT short-term 2020 benchmark \cite{vot20}, this paper uses the vot-toolkit package to evaluate the trackers. Expected average overlap (EAO), accuracy (A), and robustness (R) are employed to evaluate the performance.

\subsection{Results and Comparisons}
\subsubsection{LaSOT}
LaSOT \cite{LaSOT} is a large-scale benchmark for visual single object tracking with 1400 long sequences, which contains 14 tracking challenges. Following the Protocol \uppercase\expandafter{\romannumeral2} of LaSOT \cite{LaSOT}, this paper evaluates the performance of LightFC and LightFC-vit on the LaSOT \cite{LaSOT} test subset with 280 sequences. The results are shown in Table \ref{tab:table1}. LightFC achieves state-of-the-art performance. Compared to other lightweight conv-based trackers, LightFC outperforms FEAR-L \cite{FEAR} by 2.6 \% in AUC and 2.3 \% in precision. Compared to the attention-based lightweight tracker, LightFC achieves higher precision, and outperforms the best attention-based tracker MixFormerV2-S \cite{MixFormerV2} by 2.8 \%. LightFC-vit further outperforms MixFormerV2-S \cite{MixFormerV2} by 2.9 \% in AUC and 7.6 \% in precision. The results show that LightFC-vit and LightFC achieve competitive tracking performance in both success rate and precision.

\begin{figure}
	\centering
        \includegraphics[width=8cm]{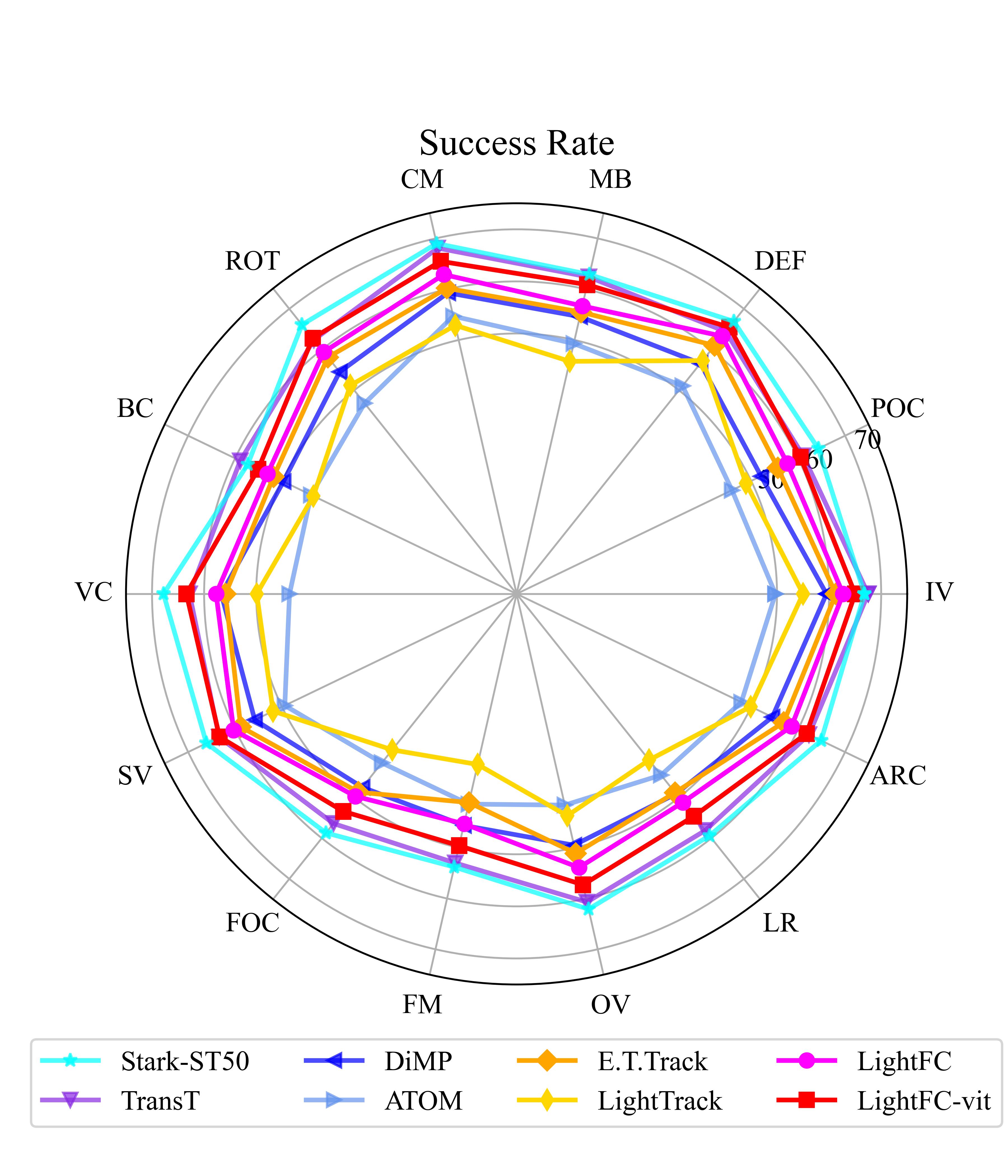}
	\caption{ifferent attribute success rate in LaSOT \cite{LaSOT}. The lightweight trackers are indicated in warmer colors. The non-lightweight trackers are indicated in colder colors.}
	\label{fig:fig5}
\end{figure}

In addition, Fig. \ref{fig:fig5} shows the AUC score on 14 LaSOT \cite{LaSOT} attributes for 8 trackers (Stark \cite{Stark}, TrDiMP \cite{TrDiMP}, LightFC, LightFC-vit, LightTrack \cite{LightTrack}, TransT \cite{TransT}, DiMP \cite{DiMP}, E.T.Track \cite{E.T.Track}, ATOM \cite{ATOM}) on 14 attributes of LaSOT \cite{LaSOT}. These attributes are {IV}: illumination variation; {POC}: partial occlusion; {DEF}: deformation; {MB}: motion blur; {CM}: camera motion; {ROT}: rotation; {BC}: background clutter; {VC}: view change; {SV}: scale variation; {FOC}: full occlusion; {FM}: fast motion; {OV}: out-of-view: {LR}: low resolution; {ARC}: aspect ratio change. Compared to LightTrack \cite{LightTrack} and E.T.Track \cite{E.T.Track}, LightFC and LightFC-vit have a competitive performance in dealing with 14 attributes. Especially in terms of FM, it exhibits better adaptability than E.T.Track \cite{E.T.Track}. Their feature representation helps them perform better in attributes like DEF, SV, and others. It is necessary to strengthen the robustness of LightFC and LightFC-vit since the success rate of the BC, FM, MB, and LR attributes is insufficient. In addition, they shorten the performance gap between lightweight trackers and advanced large-scale trackers.

\begin{table}[]\small  \centering
    \caption{Comparison of lightweight trackers on LaSOT, TrackingNet and TNL2K. The best two results are shown in \color{red}red \color{black}and \color{blue}blue \color{black}fonts.}
    \setlength{\tabcolsep}{2pt}
    \renewcommand{\arraystretch}{1.2}
    \begin{tabular}{c|c|ccc|ccc|ccc}
    \hline 
    
    &  & \multicolumn{3}{c|}{LaSOT}  & \multicolumn{3}{c|}{TrackingNet} & \multicolumn{3}{c}{TNL2K}\\ \cline{3-11} 
    \ & Source  & AUC  & P-Norm & P    & AUC       & P-Norm    & P   & AUC  & P-Norm  & P   \\ \hline
    LightTrack-Mobile\cite{LightTrack}  & CVPR2021     & 53.8 & -        & 53.7 & 72.5      & 77.9      & 69.5     & -       & -    & -   \\
    LightTrack-LargeB\cite{LightTrack}  & CVPR2021   & 55.5 & -        & 56.1 & 73.3      & 78.9      & 70.8     & - & -  & -     \\
    STARK-Lightning\cite{Stark}    & CVPR2021  & 58.6 & 69.0         & 57.9 & -         & -   & -  & -  & -    & -   \\
    FEAR-L\cite{FEAR}             & ECCV2022      & 57.9 & 68.6        & 60.9 & -         & -         & -  & -   & - & -      \\
    FEAR-XS\cite{FEAR}            & ECCV2022  & 53.5 & 64.1     & 54.5 & -   & -   & -     & -   & -      & -     \\
    E.T.Track\cite{E.T.Track}          & WACV2023     & 59.1 & -       & -    & 74.5      & 80.3      & 70.6     & -    & -      & -   \\
    
    MixFormerV2-S\cite{MixFormerV2}      & Arxiv2023    & \textbf{\color{blue}60.6} & {69.9} & {60.4} & 75.8 & 81.1  & {70.4} &  {47.2} & - & {41.8} \\ \hline
    
    LightFC            & -     & 60.5 & \textbf{\color{blue}70.2} & \textbf{\color{blue}63.2} & \textbf{\color{blue}77.0}      & \textbf{\color{blue}83.1}      & \textbf{\color{blue}74.1}     &  \textbf{\color{blue}49.8} &  \textbf{\color{blue}66.2} & \textbf{\color{blue}48.3} \\ 
    
    LightFC-vit            & -     & \textbf{\color{red}63.5} & \textbf{\color{red}74.3} & \textbf{\color{red}68.0} & \textbf{\color{red}77.8}      & \textbf{\color{red}84.4}      & \textbf{\color{red}75.3}     &  \textbf{\color{red}51.4} &  \textbf{\color{red}68.8} & \textbf{\color{red}51.2} \\ \hline
        
    \end{tabular}
    \label{tab:table1}
\end{table}

\subsubsection{TrackingNet}
TrackingNet \cite{TrackingNet} is a large-scale short-term benchmark for single object tracking that contains 511 sequences in its test subset. The trackers are evaluated on TrackingNet \cite{TrackingNet} and the results are reported in Table \ref{tab:table1}. LightFC outperforms all other lightweight trackers. The AUC, precision and norm-precision of LightFC are 1.2 \%, 2 \% and 3.7 \% higher than MixFormerV2-S \cite{MixFormerV2}, while using 5x fewer parameters (3.16 v.s. 16.04 M) and 4.6x fewer Flops (0.95 v.s. 4.4 G). The AUC, precision and norm-precision of LightFC-vit are 2 \%, 3.3 \%, 4.9 \% higher than MixFormerV2-S \cite{MixFormerV2}, while using 2.9x fewer parameters (5.5 v.s. 16.04 M) and 1.7x fewer Flops (2.48 v.s. 4.4 G). The experimental results demonstrate that LightFC and LightFC-vit outperform the other four lightweight trackers and achieve the optimal balance between performance, parameters, Flops and FPS.

\subsubsection{TNL2K}
TNL2K \cite{TNL2K} is a large-scale high-quality benchmark for single object tracking and has 700 sequences for testing. The results of LightFC are reported in Table \ref{tab:table1}. Compared to the MixFormerV2-S \cite{MixFormerV2}, LightFC improves the AUC and precision by 2.6 \% and 6.5 \%, respectively. It improves the AUC and precision for LightFC-vit by 4.2 \% and 9.4 \%, respectively. LightFC sets a new state-of-the-art benchmark on TNL2K \cite{TNL2K} dataset.

\subsubsection{OTB100}
OTB100 \cite{OTB100} is a classic object tracking benchmark with 100 short-term sequences. The AUC results are reported in Table 2. LightFC and LightFC-vit achieve an AUC score of 68.5 \% and 70.9 \%, and outperform E.T.Track \cite{E.T.Track} by 0.7 \% and 3.1 \%, respectively. Compared to non-lightweight trackers, LightFC performs comparably to DiMP \cite{DiMP}, with 8x fewer parameters required (3.2 v.s. 26.1 M). LightFC-vit achieves an AUC score 1.5 \% higher than TransT \cite{TransT} with 3x fewer parameters (5.5 v.s. 18.5 M) and 6x fewer flops (2.5 {v.s.} 16.8 G). LightFC outperforms all previous lightweight trackers with 68.5 \% AUC.

\begin{table}[h]\small \centering
    \renewcommand{\arraystretch}{1.1}
    \caption{Comparison of state-of-the-art lightweight and non lightweight trackers on OTB100, UAV123 and TC128 in terms of AUC. The best results of the lightweight tracker are shown in \color{red}red \color{black} font and the best results of the non-lightweight tracker are shown in \color{blue}blue \color{black}font.}
    \setlength{\tabcolsep}{2pt}
    \begin{tabular}{c|cccccc|ccccc}
    \hline  
        & \multicolumn{6}{c|}{\thead{non-lightweight}}               & \multicolumn{5}{c}{\thead{lightweight}}                  \\ \cline{2-12}
        
        & \thead{ATOM \\ \cite{ATOM}}  & \thead{DiMP \\ \cite{DiMP}} & \thead{SiamRCNN\\ \cite{SiamRCNN}} & \thead{Stark \\ \cite{Stark}} & \thead{KeepTrack \\ \cite{KeepTrack}} & \thead{TransT \\ \cite{TransT} } & \thead{LightTrack \\ \cite{LightTrack}} & \thead{E.T.Track \\ \cite{E.T.Track}} & \thead{MixFormerV2-S \\ \cite{MixFormerV2}} & \thead{LightFC \\ \ }  & \thead{LightFC-vit\\ \ }  \\ \hline
        
\thead{OTB100} & 66.9 & 68.4 & 70.1      & 67.3  & \textbf{\color{blue}70.9}      & 69.4   & 66.2       & 67.8      & -   & 68.5 & \textbf{\color{red}70.9}   \\
\thead{UAV123}  & 64.2 & 65.3 & 64.9      & 68.8  & \textbf{\color{blue}69.7}      & 69.1   & 62.5       & 62.3      & \textbf{\color{red}65.1}          & 64.8  & 64.9  \\
\thead{TC128}   & 59.5 & 60.8 & -         & \textbf{\color{blue}62.5}  & -         & 59.6   & 55.0         & 57.1      & -      & 61.0       & \textbf{\color{red}63.5}    \\ \hline
    \end{tabular}

    \label{tab:table2}
\end{table}

\subsubsection{UAV123}
UAV123 \cite{UAV123} is a classic benchmark for visual object tracking from aerial viewpoints. It contains a total of 123 sequences. The AUC results are reported in Table \ref{tab:table2}. LightFC and LightFC-vit surpass LightTrack \cite{LightTrack} by 2.5 \% and 2.6 \%. They perform comparably to MixFormerV2-S \cite{MixFormerV2} and SiamR-CNN \cite{SiamRCNN}. Overall, LightFC and LightFC-vit achieve competitive performance in UAV tracking tasks.

\subsubsection{TC128}
TC128 \cite{TC128} is a classic visual tracking benchmark that focuses on color information. It has 128 color sequences. As shown in Table \ref{tab:table2}, compared to LightTrack \cite{LightTrack} and E.T.Track \cite{E.T.Track}, LightFC improves the AUC by 6.0 \% and 3.9 \%, respectively. LightFC performs comparably to DiMP \cite{DiMP}. LightFC-vit achieves better AUC, being 1.0 \% higher than Stark-ST50 \cite{Stark}. LightFC and LightFC-vit demonstrate better adaptability and higher performance in color information challenges.

\subsubsection{UOT100}

UOT100 \cite{UOT100} is a typical visual underwater object tracking benchmark and contains 100 short-term underwater image sequences, which mainly address the challenge of underwater image distortion in different scenes. Since there are no official reports on the results of the UOT100 \cite{UOT100} for lightweight trackers, this paper compares LightFC with non-lightweight trackers. Table \ref{tab:table3} presents that LightFC achieves better AUC (61.6 \%) and better precision (52.2 \%), both of which are 1.0 \% higher than those of KeepTrack \cite{KeepTrack}. LightFC-vit performs comparably to TransT \cite{TransT}. In addition, the precision score of LightFC-vit is 2.4 \% higher than that of TransT \cite{TransT}. In the challenge of underwater image degradation, LightFC and LightFC-vit achieve competitive performance with much fewer parameters and Flops.

\begin{table}[h]\small \centering
    \renewcommand{\arraystretch}{1.1}
    \caption{Comparison of state-of-the-art lightweight and non-lightweight trackers on UOT100. The best results of lightweight tracker are shown in \color{red}red \color{black}font and the best results of non-lightweight tracker are shown in \color{blue}blue \color{black}fonts.}
    \begin{tabular}{c|ccccccc|cc}
    \hline
      & \thead{SiamFC \\ \cite{SiamFC}} & \thead{SiamRPN \\ \cite{SiamRPN}}& \thead{SiamCAR \\ \cite{SiamCAR}} & \thead{DiMP \\ \cite{DiMP}}& \thead{Stark \\ \cite{Stark}} & \thead{KeepTrack \\ \cite{KeepTrack}} & \thead{TransT \\ \cite{TransT}} & \thead{LightFC \\ \ } & \thead{LightFC-vit \\ \ } \\ \hline
AUC   & 43.8   & 59.7    & 53.6    & 59.8 & \textbf{\color{blue}66.3}  & 60.6      & 63.8   &61.6  & \textbf{\color{red}63.9}    \\
P-Norm & 53.4   & 74.8    & 69.4    & 75.4 & \textbf{\color{blue}81.6}  & 78.1     & 79.9   &75.8  & \textbf{\color{red}78.2}    \\
P     & 30.4   & 48.7    & 46.0      & 48.9 & \textbf{\color{blue}57.9}  & 51.2     & 56.3  &52.2  & \textbf{\color{red}53.6}    \\ \hline
    \end{tabular}
    \label{tab:table3}
\end{table}

\begin{figure}
	\centering
        \includegraphics[width=8cm]{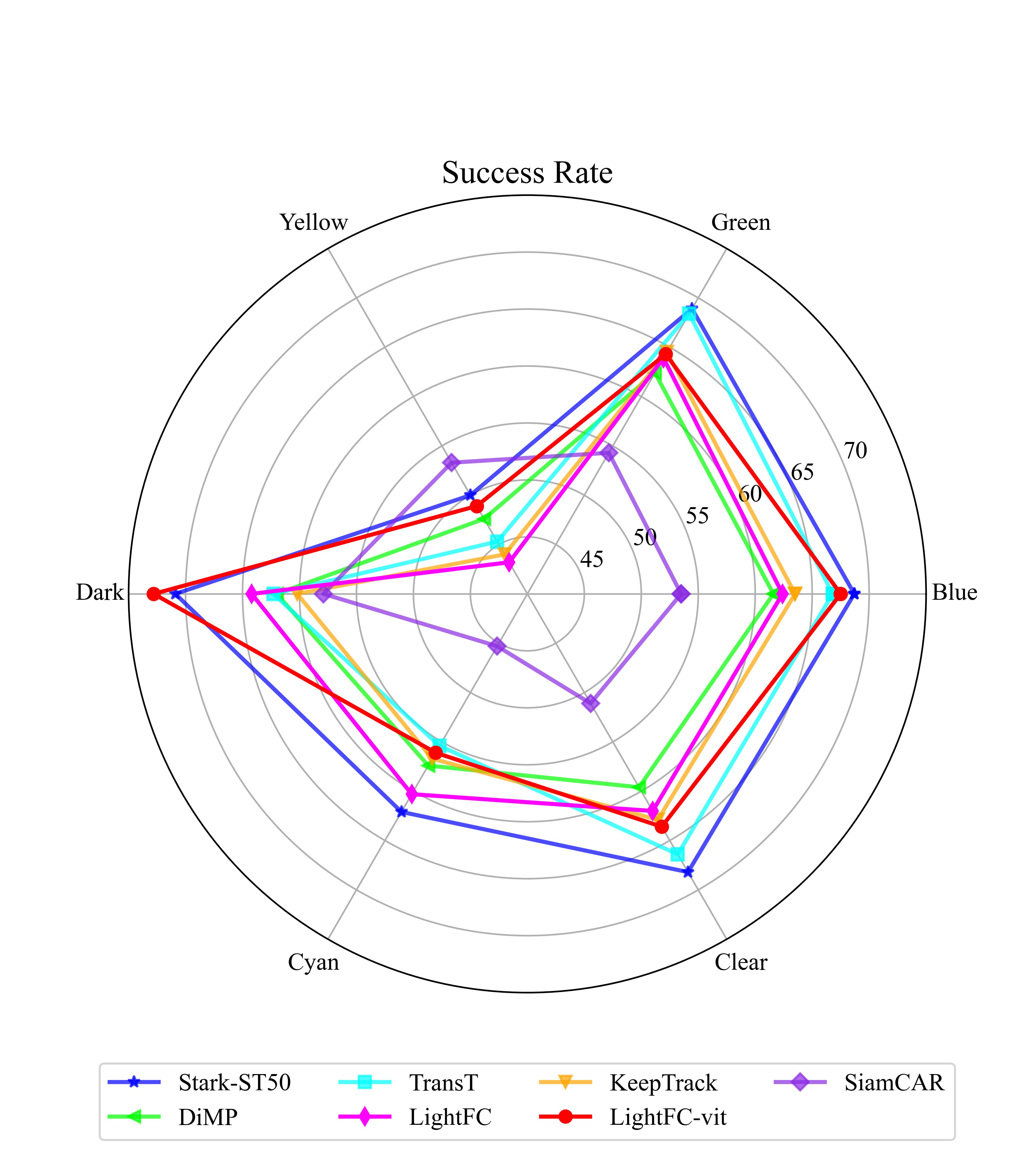}
	\caption{Different color distortion attribute success rate on UOT100 (divided by \cite{UStark}).}
	\label{fig:fig6}
\end{figure}

Fig. \ref{fig:fig6} presents the AUC score of the trackers of the best six in Table \ref{tab:table3} on 6 color distortion attributes of UOT100 \cite{UOT100}. LightFC performs competitively in dealing with dark and cyan color distortion. However, performance decreases while dealing with yellow distortion. LightFC-vit significantly improves LightFC’s performance in Dark, Blue, and Yellow color distortion. However, it has shortcomings in dealing with yellow and cyan color distortion.

\subsubsection{UTB180}
UTB180 \cite{UTB180} is a high-quality underwater object tracking benchmark with 180 short-term sequences. It reflects the challenges in tracking underwater targets, such as scale variation, occlusion, similar objects and so on. For the same reason as UOT100 [43], this paper compares LightFC with non-lightweight trackers. As shown in Table \ref{tab:table4}, Compared to KeepTrack \cite{KeepTrack}, LightFC improves the AUC by 0.3 \% and precision by 1.1 \%, respectively. Compared to TransT \cite{TransT}, LightFC-vit improves the AUC by 1.4 \% and norm-precision by 3.5 \%, respectively. In typical underwater tracking tasks, LightFC also achieves competitive performance with much fewer parameters and Flops.

\begin{table}[h]\small \centering
    \renewcommand{\arraystretch}{1.1}
    \caption{Comparison of state-of-the-art lightweight and non lightweight trackers on UTB180. The best results of lightweight tracker are shown in \color{red}red \color{black}font and the best results of non-lightweight tracker are shown in \color{blue}blue \color{black}fonts.}
    \begin{tabular}{c|ccccccc|cc}
    \hline
      & \thead{SiamFC \\ \cite{SiamFC}} & \thead{SiamRPN \\ \cite{SiamRPN}}& \thead{SiamCAR \\ \cite{SiamCAR}} & \thead{DiMP \\ \cite{DiMP}}& \thead{Stark \\ \cite{Stark}} & \thead{KeepTrack \\ \cite{KeepTrack}} & \thead{TransT \\ \cite{TransT}} & \thead{LightFC \\ \ } & \thead{LightFC-vit \\ \ }\\ \hline
AUC   & 35.0   & 53.4    & 49.8    & 50.5 & 55.9  & 54.6      & \textbf{\color{blue}57.5} &54.9  & \textbf{\color{red}58.9} \\
P-Norm & 41.2   & 63.5    & 60.1    & 58.5 & 64.6  & 64.1      & \textbf{\color{blue}66.1}&63.7  & \textbf{\color{red}69.6} \\
P     & 22.8   & 41.9    & 40.8    & 37.5 & 48.5  & 43.5      & \textbf{\color{blue}50.3} &44.6  & \textbf{\color{red}50.0} \\ \hline
    \end{tabular}
    \label{tab:table4}
\end{table}

\begin{figure}
	\centering
        \includegraphics[width=8cm]{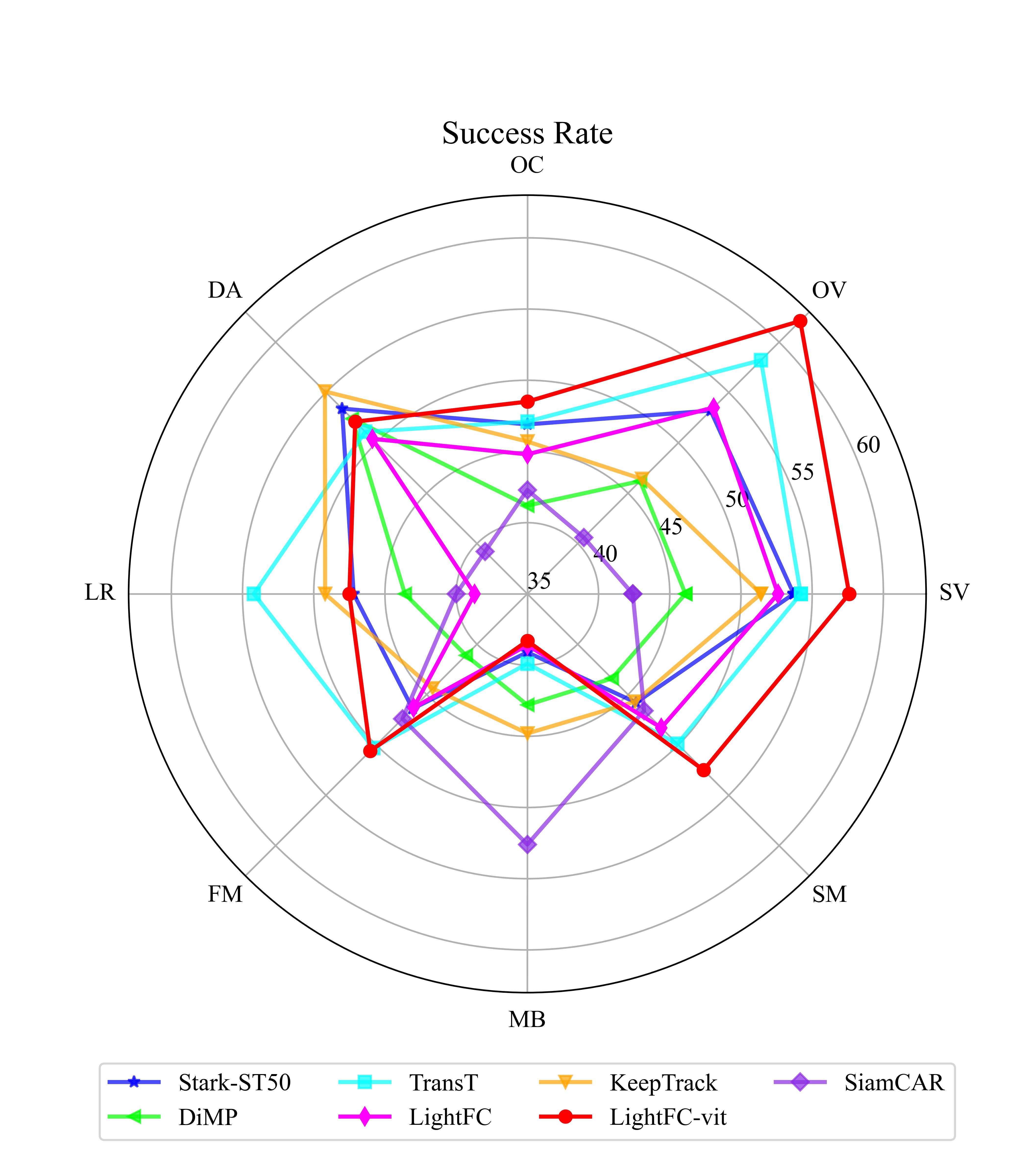}
	\caption{Different attribute success rate on UTB180 (officially divided).}
	\label{fig:fig7}
\end{figure}

Fig. \ref{fig:fig7} presents the AUC score of the trackers of the best six from Table \ref{tab:table4} for 8 attributes of UTB180 \cite{UTB180}. These attributes are SV: scale variation; OV: out-of-view; OC: occlusion (both partial occlusion and full occlusion); DA: deformation; LR: low resolution; FM: fast motion; MB: motion blur; SM: similar object. Both LightFC and LightFC-vit have insufficient success rate in low resolution and motion blur attributes. This indicates that they have shortcomings in discriminating low-frequency information in underwater images.

\subsubsection{VOT20-ST}
VOT-ST2020 \cite{vot20} benchmark has 60 challenging sequences. EAO, A and R are used to evaluate the performance of trackers. Table 4 shows that LightFC performs better than existing lightweight trackers, surpassing LightTrack \cite{LightTrack} and MixFormerV2-S \cite{MixFormerV2} by 3.1 \% and 1.5 \% in terms of EAO, respectively, and achieves superior performance compared to existing lightweight trackers. LightFC-vit outperforms other lightweight trackers with its 0.292 EAO, 0.466 A, and 0.768 R. Overall, LightFC and LightFC-vit achieve state-of-the-art performance for lightweight trackers in the VOT-ST2020 challenge.

\begin{table}[h]\small \centering
    \renewcommand{\arraystretch}{1.1}
    \setlength{\tabcolsep}{3pt}
    \caption{Comparison of state-of-the-art lightweight and non lightweight trackers on VOT-ST2020. The best results of lightweight tracker are shown in \color{red}red \color{black} font and the best results of non-lightweight tracker are shown in \color{blue}blue \color{black}font.}
    \begin{tabular}{c|ccccc|ccccc}
    \hline
        & \multicolumn{5}{c|}{Non-lightweight}            & \multicolumn{5}{c}{lightweight}                  \\ \cline{2-11} 
        & \thead{SiamFC \\ \cite{SiamFC}} &  \thead{ATOM \\ \cite{ATOM}} &  \thead{DiMP \\ \cite{DiMP}} & \thead{ Stark-S50 \\ \cite{Stark}} & \thead{ Stark-ST50 \\ \cite{Stark}}&  \thead{LightTrack \\ \cite{LightTrack}}&  \thead{ E.T.Track0 \\ \cite{E.T.Track}}& \thead{ MixFormerV2-S \\ \cite{MixFormerV2}}& \thead{LightFC \\ \ } &\thead{LightFC-vit \\ \ } \\ \hline
    EAO & 0.179  & 0.271 & 0.274 & 0.280     &\textbf{\color{blue}0.308}      & 0.242      & 0.267     & 0.258         & 0.273 & \textbf{\color{red}0.292}  \\
    A   & 0.418  & 0.462 & 0.457 & 0.477     & \textbf{\color{blue}0.478}      & 0.422      & 0.432     & -             &  0.459  &\textbf{\color{red}0.466}   \\
    R   & 0.502  & 0.734 & 0.74  & 0.728     & \textbf{\color{blue}0.799}      & 0.689      & 0.741     & -    & 0.723 & \textbf{\color{red}0.768}  \\ \hline
    \end{tabular}
    \label{tab:table5}
\end{table}

\subsubsection{Speed}
Table \ref{tab:table6} presents a comparison of the parameters, Flops, and FPS of LightFC and other lightweight trackers on different CPU and GPU configurations. The results show that LightFC achieves state-of-the-art speeds of 72 FPS, 55 FPS and 129 FPS on the Intel I9-12900KF CPU, AMD Ryzen7 4800H CPU and GTX1050Ti GPU, respectively. LightFC achieves competitive speeds at 284 FPS on GPU GTX 3090Ti. Moreover, LightFC exhibits good adaptability in terms of speed in different devices, outperforming other trackers. For instance, MixFormerV2-S \cite{MixFormerV2} boasts the highest speed (584 FPS) on the GTX 3090Ti GPU, but its speed drops on other devices. This problem is also evident in FEAR-XS \cite{FEAR} and E.T.Track \cite{E.T.Track}. LightTrack \cite{LightTrack} shows the fastest performance (95 FPS) with the I9-12900KF CPU, while its speed on GPUs is not fast. Only the speed changes of LightFC are consistent on these devices. Although LightFC-vit runs slower than LightFC, it still achieves competitive speed in GPUs.

\begin{table}[h]\small \centering
    \renewcommand{\arraystretch}{1.1}
    \setlength{\tabcolsep}{2pt}
    \caption{Run-time speed on different device. The best two results are shown in \color{red}red \color{black}and \color{blue}blue \color{black}fonts.}
    \begin{tabular}{c|cc|cccc}
    \hline
    \  & \  & \  & \multicolumn{4}{c}{FPS}    \\ \cline{4-7} 
Tracker                  & Param /M                  & Flops /G                  & \multicolumn{2}{c|}{CPU}                            & \multicolumn{2}{c}{GPU} \\ \cline{4-7} 
                         &                        &                        & Intel I9-12900KF & \multicolumn{1}{c|}{AMD R7 4800H} & GTX 3090Ti  & GTX 1050Ti  \\ \hline
LightTrack-Mobile\cite{LightTrack}    & 1.97  & 0.54                   & \textbf{\color{red}95}         & \multicolumn{1}{c|}{19}                & 119        & 61         \\
FEAR-XS \cite{FEAR}              & 1.37  & 0.48                   & 40         & \multicolumn{1}{c|}{\textbf{\color{blue}33}}                & \textbf{\color{blue}323}        & 117        \\
E.T.Track \cite{E.T.Track}       & 6.98  & 1.56                   & 20         & \multicolumn{1}{c|}{16}                & 80         & 39         \\
MixFormerV2-S \cite{MixFormerV2} & 16.04  & 4.40                    & 32         & \multicolumn{1}{c|}{20}                & \textbf{\color{red}584}        & \textbf{\color{blue}126}        \\ \hline
LightFC                  & 3.16                   & 0.95                   & \textbf{\color{blue}72}         & \multicolumn{1}{c|}{\textbf{\color{red}55}}                & 284        & \textbf{\color{red}129}        \\ 
LightFC-vit              & 5.50                   & 2.48                   & 34        & \multicolumn{1}{c|}{10}               & 202        & 98        \\ \hline

\end{tabular}
\label{tab:table6}
\end{table}

\subsection{Ablation and Analysis} \label{sec:subsection4.3}
To investigate the individual influence of each component and establish the optimal module design, this work executes several ablation experiments on LightFC and reports the results for AUC, Precision and Norm-Precision on LaSOT \cite{LaSOT} and TNL2K \cite{TNL2K}.

\subsubsection{Backbone} \label{sec:subsection4.3.1}
The performance of various efficient backbone networks is evaluated in the tracking pipeline and the results for the optimal selection are reported in Table \ref{tab:table7}. The output feature stride of each backbone is set at 16. After thorough comparison, this work selects MobileNetV2 \cite{MobilNetV2} and TinyViT \cite{tinyvit} as the backbone for LightFC and LightFC-vit due to their superior performance.

\begin{table}[h]\small \centering
    \renewcommand{\arraystretch}{1.1}
    \setlength{\tabcolsep}{4pt}
    \caption{Ablation of the different backbone of LightFC.}
    \begin{tabular}{c|ccc|ccc}
    \hline
    \  & \multicolumn{3}{c|}{LaSOT} & \multicolumn{3}{c}{TNL2K} \\ \cline{2-7} 
                  & AUC     & P-Norm   & P      & AUC    & P-Norm   & P      \\ \hline

ResNet18\cite{resnet}       & 60.0    & 69.5    & 62.2     & 48.0   & 63.6   & 46.3   \\
ShuffleNetV2 \cite{ShuffleNetV2}       & 57.8    & 66.9    & 59.4     & 47.5   & 44.8    & 63.1  \\
MobilenetV2\cite{MobilNetV2}       & \textbf{60.4}    & \textbf{70.0}      & \textbf{63.1}   & \textbf{49.8}   & \textbf{66.2}    & \textbf{48.3}   \\
MobilenetV3\cite{MobilNetV3}       & 56.6    & 65.2    & 57.0     & 46.1   & 61.1    & 41.9   \\
LT-Mobile\cite{LightTrack}         & 58.9    & 67.7    & 61.2   & 48.4   & 63.3    & 45.2   \\
EfficientNet \cite{EfficientNet}   & 59.3    & 69.2    & 62.0   & 49.4   & 66.5    & 47.9   \\
LCNet\cite{pplcnet}                & 58.5    & 68.7    & 61.1   & 48.3   & 64.9    & 45.9   \\
MobileOne\cite{MobileOne}          & 51.5    & 60.8    & 50.7   & 43.4   & 58.0      & 37.4   \\ \hline
TinyViT\cite{tinyvit}              & \textbf{63.5}    & \textbf{74.3}      & \textbf{68.0}   & \textbf{51.4}   & \textbf{68.8}    & \textbf{51.2}   \\
LightViT\cite{lightvit}            & 59.6    & 69.0    & 62.2   & 49.4   & 64.9    & 47.5   \\
MobileFormer \cite{MobileFormer}   & 56.3    & 65.0    & 56.9   & 46.0   & 61.2    & 42.6   \\
MiniDeiT \cite{MiniDeiT}           & 60.8    & 70.3    & 63.3   & 49.8   & 65.7    & 47.9   \\ \hline
    \end{tabular}
    \label{tab:table7}
\end{table}

\subsubsection{ECM}
This work evaluates the contributions of the three factors presented in Section \ref{sec:subsection3.2} to ECM and explores the function of each factor.

Firstly, this paper evaluates the impact of incorporating nonlinear blocks into the ECM. The results are presented in Table \ref{tab:table8}. An improvement of 2.1 \% and 1.5 \% in terms of AUC, 2.4 \% and 1.8 \% in terms of norm-precision on LaSOT \cite{LaSOT} and TNL2K \cite{TNL2K} is obtained by the contribution of the SCF unit and the IAB. Overall, improving feature representation and enhancing model nonlinearity can effectively enhance the performance of ECM.

\begin{table}[h]\small \centering
    \renewcommand{\arraystretch}{1.1}
    \setlength{\tabcolsep}{4pt}
    \caption{Ablation of the nonlinear blocks of ECM.}
    \begin{tabular}{c|cc|ccc|ccc}
    \hline
    \                      & SCF & IAB& \multicolumn{3}{c|}{LaSOT} & \multicolumn{3}{c}{TNL2K} \\ \cline{4-9} 
                                      &                      &                      & AUC     & P-Norm  & P      & AUC    & P-Norm  & P      \\ \hline
                                      & \Checkmark           & \Checkmark           & \textbf{60.5} & \textbf{70.2} & \textbf{63.2} & \textbf{49.8} & \textbf{66.2}  & \textbf{48.3}   \\
    \ nonlinear blocks                & \Checkmark           &                      & 58.8    & 68.2    & 61.4   & 49.1   & 65.3    & 47.6   \\
                                      &                      & \Checkmark           & 59.2    & 68.8    & 61.9   & 49.2   & 65.4    & 47.8   \\
                                      &                      &                      & 58.4    & 67.8    & 61.0   & 48.3   & 64.4    & 46.7   \\ \hline
    \end{tabular}
    \label{tab:table8}
\end{table}

After analyzing the performance change before and after adding nonlinear blocks in Table \ref{tab:table8}, it is clear that using only the LightTrack \cite{LightTrack} and E.T.Track \cite{E.T.Track} feature fusion modules (which only use pixel-wise correlation and the SE \cite{SE}] module) results in an insufficient expressiveness of the fused features. SCF implements the fusion of local spatial features and channel information in each layer of the fused features. Furthermore, the inverted activation of features in IAB effectively enhances the model’s nonlinearity.

Secondly, this paper evaluates the impact of different skip-connection on tracker performance. As shown in Table \ref{tab:table9}, the skip-connections are critical for ECM in both SCF and IAB. They improve the performance by 2.2 \% and 1.0 \% in terms of AUC, and 2.3 \% and 1.6 \% in terms of precision, respectively. 

\begin{table}[h]\small \centering
    \renewcommand{\arraystretch}{1.1}
    \setlength{\tabcolsep}{4pt}
    \caption{Ablation of the skip-connections of different nonlinear block of ECM.}
    \begin{tabular}{c|cc|ccc|ccc}
    \hline
    \                      & SCF & IAB & \multicolumn{3}{c|}{LaSOT} & \multicolumn{3}{c}{TNL2K} \\ \cline{4-9} 
                  &                      &                      & AUC     & P-Norm     & P      & AUC    &  P-Norm  & P      \\ \hline
                  & \Checkmark           & \Checkmark           & \textbf{60.5} & \textbf{70.2} & \textbf{63.2} & \textbf{49.8} & \textbf{66.2}  & \textbf{48.3}   \\
skip-connection   & \Checkmark           &                      & 59.5    & 69.3    & 62.4   & 49.5   & 66.0      & 48.0     \\
                  &                      & \Checkmark           & 58.6    & 68.1    & 61.1   & 49.4   & 65.5      & 47.7   \\
                  &                      &                      & 58.3    & 67.7    & 60.9   & 48.8   & 64.9      & 47.3   \\ \hline
    \end{tabular}
    \label{tab:table9}
\end{table}

Thirdly, the prediction head is divided into two branches: CLS (the classification branch of ERH) and BOX (the offset and size branch). The impact of reusing search area features on performance is explored in both branches. The results in Table \ref{tab:table10}  indicates that the reuse of search range features leads to performance improvements in each branch. Feature reuse in both the CLS and BOX branches results in a combined improvement of 2.3 \% and 1.6 \% for AUC, and 2.5 \% and 2.4 \% for PNORM improvement on LaSOT \cite{LaSOT} and TNL2K \cite{TNL2K}, respectively. 

\begin{table}[h]\small \centering
    \renewcommand{\arraystretch}{1.1}
    \setlength{\tabcolsep}{4pt}
    \caption{Ablation of the features reuse of different branch.}
    \begin{tabular}{c|cc|ccc|ccc}
    \hline
    \             & \ CLS                &  BOX                 & \multicolumn{3}{c|}{LaSOT} & \multicolumn{3}{c}{TNL2K} \\ \cline{4-9} 
                  &                      &                      & AUC     & P-Norm  & P      & AUC    & P-Norm  & P      \\ \hline
concatenate       & \Checkmark           & \Checkmark           & \textbf{60.5} & \textbf{70.2} & \textbf{63.2} & \textbf{49.8} & \textbf{66.2}  & \textbf{48.3}   \\
                  & \Checkmark           &                      & 59.1    & 61.8    & 68.8   & 48.6   & 64.6    & 46.5   \\
                  &                      & \Checkmark           & 59.7    & 69.2    & 62.4   & 49.4   & 65.5    & 47.9   \\
                  &                      &                      & 58.2    & 67.7    & 60.8   & 48.2   & 63.8    & 46.4   \\ \hline
    \end{tabular}
    \label{tab:table10}
\end{table}

To better reuse search area features, this work investigates the performance of the tracker when substituting the concatenation operation for the addition operation.  Table \ref{tab:table11} presents that the concatenation operation performs better than the addition operation.

\begin{table}[h]\small \centering
    \renewcommand{\arraystretch}{1.1}
    \setlength{\tabcolsep}{4pt}
    \caption{Ablation of the different method for reusing search area features.}
    \begin{tabular}{c|ccc|ccc}
    \hline
    \             & \multicolumn{3}{c|}{LaSOT} & \multicolumn{3}{c}{TNL2K} \\ \cline{2-7} 
                  & AUC     & P-Norm  & P      & AUC    & P-Norm  & P      \\ \hline
concat            & \textbf{60.5} & \textbf{70.2} & \textbf{63.2} & \textbf{49.8} & \textbf{66.2}  & \textbf{48.3}   \\
add               & 59.3    & 68.7    & 61.6   & 48.6   & 64.0      & 46.2   \\ \hline
    \end{tabular}
    \label{tab:table11}
\end{table}

\subsubsection{ERH}
\begin{figure}
	\centering
        \includegraphics[width=14cm]{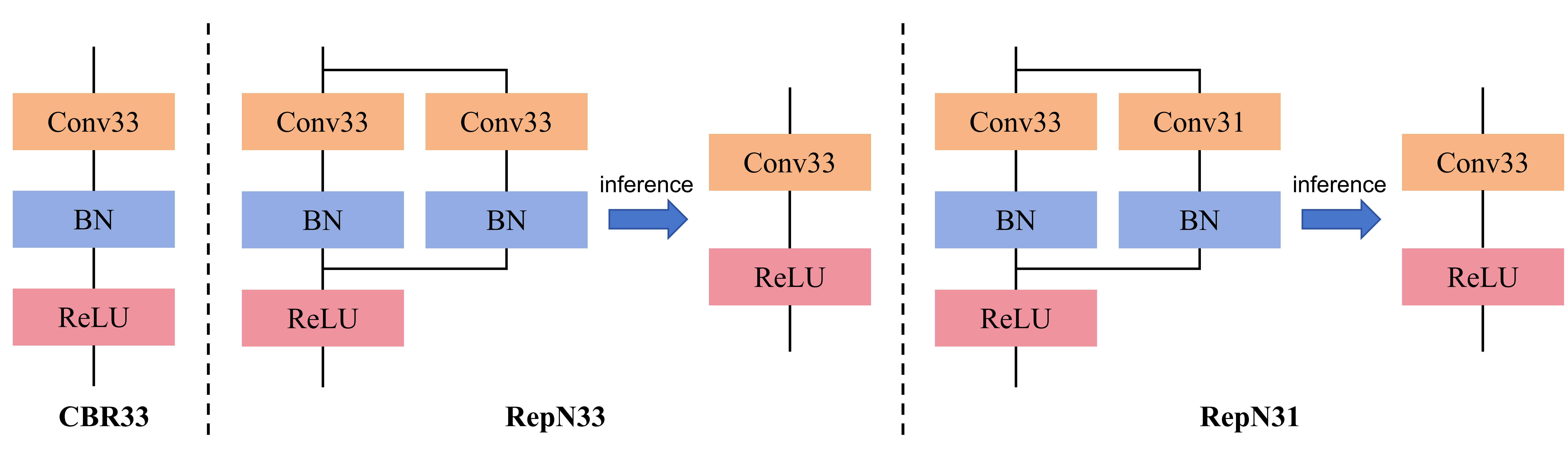}
	\caption{The structure of Conv33, RepN33 and RepN31 in ERH.}
	\label{fig:fig8}
\end{figure}

The center-head \cite{centerhead} branch is divided into two stages. The first stage consists of $F_{33}^{1}$,while the second stage consists of $F_{33}^{1},i\in \{2,3,4\}$. The impact of reparameterization and the characteristics of the feature flow are evaluated. The reparameterization models are shown in Fig. \ref{fig:fig8}. Table \ref{tab:table12} reports the results and the phenomenon can be summarized in two key aspects.

1.Using the reparameterization technique in stage 1 can effectively improve model performance, while it decreases model performance in stage 2.

2.RepN33 block always do better than RepN31 block.

For the first point, RepN33 does better than CBR33 in the first stage, which demonstrates the feature representation bottleneck of stage1 of the standard center head \cite{centerhead} branch. Therefore, using two 3x3 convolutional kernels to reparametrize and enrich the expression of new features is identified as a key factor for the success of ERH. Additionally, during the second stage of feature dimension reduction, it is observed that convolutional kernels can focus on compressing and extracting crucial information. Redundant convolutional kernels in reparameterization may reintroduce interference information that has already been filtered out into the features, which can affect the model’s comprehension of significant features.

\begin{table}[h]\small \centering
    \renewcommand{\arraystretch}{1.1}
    \setlength{\tabcolsep}{4pt}
    \caption{Ablation of the different reparameterization kernel of Center-Head.}
    \begin{tabular}{c|cc|ccc|ccc}
    \hline
    \      &  Stage1 &  Stage2 & \multicolumn{3}{c|}{LaSOT} & \multicolumn{3}{c}{TNL2K} \\ \cline{4-9} 
                               &                         &                         & AUC     & P-Norm  & P      & AUC   & P-Norm   & P      \\ \hline
                               & RepN33                  & RepN33                  & 59.7    & 69.1    & 62.2   & 49.3   & 65.7    & 47.6   \\
                               & RepN33                  & RepN31                  & 59.2    & 68.6    & 61.5   & 49.5   & 65.7    & 48.3   \\
                               & RepN31                  & RepN33                  & 58.5    & 67.9    & 61.1   & 49.6   & 66.1    & 48.6   \\
                               & RepN31                  & RepN31                  & 58.8    & 68.0    & 61.0   & 49.5   & 65.5    & 48.1   \\ \cline{2-9} 
     RepN-Conv       & CBR33                  & RepN33                  & 58.5    & 67.7    & 60.3   & 48.6   & 64.3    & 46.3   \\
                               & CBR33                  & RepN31                  & 58.2    & 67.3    & 59.9   & 48.6   & 64.1    & 46.7   \\ \cline{2-9} 
                               & CBR33                  & CBR33                  & 59.6    & 69.2    & 62.1   & 49.3   & 65.1    & 46.8   \\ \cline{2-9} 
                               & RepN33                  & CBR33                  & \textbf{60.5} & \textbf{70.2} & \textbf{63.2} & \textbf{49.8} & \textbf{66.2}  & \textbf{48.3}   \\ 
                               & RepN31                  & CBR33                  & 59.2    & 68.4    & 61.5   & 49.7   & 65.9    & 48.1   \\ \hline
    \end{tabular}
    \label{tab:table12}
\end{table}

\begin{table}[h]\small \centering
    \renewcommand{\arraystretch}{1.1}
    \setlength{\tabcolsep}{4pt}
    \caption{Ablation of the different component of ERH.}
    \begin{tabular}{c|ccc|ccc|ccc}
    \hline
                         & Stage1      & -  & Stage2 & \multicolumn{3}{c|}{LaSOT} & \multicolumn{3}{c}{TNL2K} \\ \cline{2-10} 
                         & RepN33      & SE & Conv   & AUC             & P-Norm  & P      & AUC    & P-Norm   & P      \\ \hline
    ERH      & \Checkmark  & \Checkmark  & \Checkmark      & \textbf{60.5}  & \textbf{70.2} & \textbf{63.2} & \textbf{49.8} & \textbf{66.2}  & \textbf{48.3}   \\
                         & \Checkmark  &             & \Checkmark      & 59.6    & 69.2    & 61.9   & 49.4   & 65.3    & 47.5   \\
                         &             & \Checkmark  & \Checkmark      & 59.5    & 69.0    & 61.9   & 49.7   & 65.9    & 48.6   \\
                         &             &             & \Checkmark      & 59.3    & 68.7    & 61.3   & 49.1   & 64.9    & 46.8   \\ \hline
    \end{tabular}
    \label{tab:table13}
\end{table}

For the second point, the 1x1 convolution kernel is mostly focused on learning channel features within a single spatial point, while the 3x3 convolutional kernel can learn other local spatial features in addition to channel features. This may indicate that center-head \cite{centerhead} requires a stronger representation of spatial features rather than channel features.

Then each component’s contribution to ERH is measured. Table \ref{tab:table13} presents the results of the ablation experiment for ERH. Implementing the RepN33 block and the SE \cite{SE} module results in performance improvements. In LaSOT \cite{LaSOT} and TNL2K \cite{TNL2K}, ERH surpasses center-head \cite{centerhead}, with increases of 1.2 \% and 0.7 \% for AUC and 1.9 \% and 1.5 \% for precision score, respectively.

\subsubsection{IOU}
To fully optimize the tracking pipeline of the model, this paper evaluates several IoU loss functions and presents the results in Table \ref{tab:table14}. WioU \cite{wiou} achieves an average improvement of 0.2 \% in both AUC and precision on LightFC compared to the widely used GIoU \cite{giou}, outperforming all other IoU loss functions. Consequently, WIoU \cite{wiou} is selected as the IoU loss function for the trackers.

\begin{table}[h]\small \centering
    \renewcommand{\arraystretch}{1.1}
    \setlength{\tabcolsep}{4pt}
    \caption{Ablation experiments on different IoU function.}
    \begin{tabular}{c|ccc|ccc}
    \hline
              & \multicolumn{3}{c|}{LaSOT} & \multicolumn{3}{c}{TNL2K} \\ \cline{2-7} 
    Loss      & AUC     & PNORM   & P      & AUC    & PNORM   & P      \\ \hline
    GIoU \cite{giou} & 60.2    & 69.8    & 62.8   & 49.6   & 65.6    & 46.9   \\
    CIoU \cite{ciou} & 60.4    & 70.0    & 63.0   & 49.6   & 66.0    & 48.2   \\
    EIoU \cite{eiou} & 58.7    & 68.4    & 61.6   & 48.9   & 65.5    & 47.7   \\
    SIoU \cite{siou} & 59.7    & 69.1    & 62.2   & 49.3   & 65.6    & 47.4   \\
    WIoU \cite{wiou} & \textbf{60.5} & \textbf{70.2} & \textbf{63.2} & \textbf{49.8} & \textbf{66.2}  & \textbf{48.3}   \\ \hline
    \end{tabular}
    \label{tab:table14}
\end{table}

\subsection{Hyperparameter Analysis}
\subsubsection{Input Size}
Table \ref{tab:table15} reports results from the evaluation of the model’s parameters, flops, and performance for different input sizes. Performance is degraded in LightFC192 and LightFC224 because smaller input sizes limit the model’s ability to represent key features by reducing the spatial resolution of the feature maps. However, LightFC384 and LightFC448 show the same performance degradation. Although a larger input image increases the spatial feature resolution, it dilutes the key features contained in each patch of the feature map. In pixel-wise correlation, each patch in the template feature interacts globally with the search area feature, which weakens the representation of the fused features as fewer key features make them more susceptible to interference and inaccurate responses. This may account for the decrease in model performance when the input size is increased from 256 to 384. However, when the input size is increased to 448, the performance of LightFC448 instead outperforms LightFC384, suggesting that while increasing the model’s input size by itself does not necessarily improve the performance, higher feature resolution may help do so.

\begin{table}[h]\small \centering
    \renewcommand{\arraystretch}{1.1}
    \setlength{\tabcolsep}{4pt}
    \caption{Ablation of the different input sizes of LightFC. The size of template is half of the search area.}
    \begin{tabular}{c|cc|ccc|ccc}
    \hline
    \ Input Size of Search Area & Params /M & Flops /G & \multicolumn{3}{c|}{LaSOT} & \multicolumn{3}{c}{TNL2K} \\ \cline{4-9} 
    \     &           &        & AUC     & P-Norm  & P      & AUC    & P-Norm  & P      \\ \hline
    \ 192  & 3.15    & 0.53    & 55.7    & 65.7    & 56.5   & 45.3   & 41.4    & 60.7   \\
    \ 224  & 3.15    & 0.72    & 58.1    & 68.1    & 59.7   & 46.9   & 43.8    & 62.7   \\
    \ 256  & \textbf{3.16}    & \textbf{0.95}    & \textbf{60.4}    & \textbf{70.2}    & \textbf{63.1}   & \textbf{49.8}   & \textbf{48.3}    & \textbf{66.2}   \\
    \ 384  & 3.28    & 2.18    & 58.3    & 66.6    & 60.3   & 47.9   & 45.7    & 62.3   \\
    \ 448  & 3.39    & 3.03    & 59.7    & 67.1    & 61.4   & 49.7   & 47.4    & 64.0   \\ \hline 

    \end{tabular}
    \label{tab:table15}
\end{table}

\subsubsection{Channel Expansion Rate of IAB}
To improve the feature representation, this article analyzes the impact of channel expansion rate $\alpha$ of IAB on model performance. Table \ref{tab:table16} shows both smaller and larger channel expansion rates in IAB impair the performance of the model. Overall, $\alpha=2$ is the optimal channel expansion rate setting.

\begin{table}[h]\small \centering
    \renewcommand{\arraystretch}{1.1}
    \setlength{\tabcolsep}{4pt}
    \caption{Ablation of the channel expansion rate of IAB}
    \begin{tabular}{c|ccc|ccc}
    \hline
    Channel expansion rate of IAB          & \multicolumn{3}{c|}{LaSOT} & \multicolumn{3}{c}{TNL2K} \\ \cline{2-7} 
           & AUC     & PNORM   & P      & AUC    & PNORM   & P      \\ \hline
    0.5   & 57.6    & 70.0    & 60.3   & 47.1   & 62.4    & 44.3   \\
    1     & 58.3    & 68.0    & 60.7   & 47.7   & 63.3    & 45.1   \\
    2     & \textbf{60.4} & \textbf{70.2} & \textbf{63.1} & \textbf{49.8} & \textbf{66.2}  & \textbf{48.3}   \\
    3     & 58.4    & 67.8    & 60.7   & 47.6   & 63.3    & 45.5   \\
    4     & 58.5    & 68.1    & 61.1   & 47.9   & 63.6    & 45.8   \\ \hline
    \end{tabular}
    \label{tab:table16}
\end{table}

\subsubsection{Output Feature Channel of ERH}
To explore whether increasing feature channels can improve feature representation bottlenecks, this paper conducts ablation experiments on the output feature channel $C_{head}$. As shown in Table \ref{tab:table17}, 256 is the optimal choice for $C_{head}$ . Increasing the number of output feature channels alone does not improve feature representation. On the contrary, it significantly reduces the performance of the model.

\begin{table}[h]\small \centering
    \renewcommand{\arraystretch}{1.1}
    \setlength{\tabcolsep}{4pt}
    \caption{Ablation of the different output channels of stage 1 of ERH}
    \begin{tabular}{c|ccc|ccc}
    \hline
    Output feature channels of stage 1  & \multicolumn{3}{c|}{LaSOT} & \multicolumn{3}{c}{TNL2K} \\ \cline{2-7} 
            & AUC     & PNORM   & P      & AUC    & PNORM   & P      \\ \hline
    192     & 57.0    & 66.3    & 59.3   & 46.2   & 61.7    & 43.6   \\
    256     & \textbf{60.4} & \textbf{70.2} & \textbf{63.1} & \textbf{49.8} & \textbf{66.2}  & \textbf{48.3}   \\
    384     & 58.7    & 68.4    & 61.2   & 68.4   & 63.6    & 45.4   \\
    448     & 57.0    & 66.0    & 59.1   & 47.5   & 63.1    & 45.2   \\
    512     & 56.1    & 65.4    & 58.5   & 47.0   & 62.6    & 44.9   \\ \hline
    \end{tabular}
    \label{tab:table17}
\end{table}

\subsection{Visualization}
\subsubsection{Heatmap}
\begin{figure}
	\centering
        \includegraphics[width=10cm]{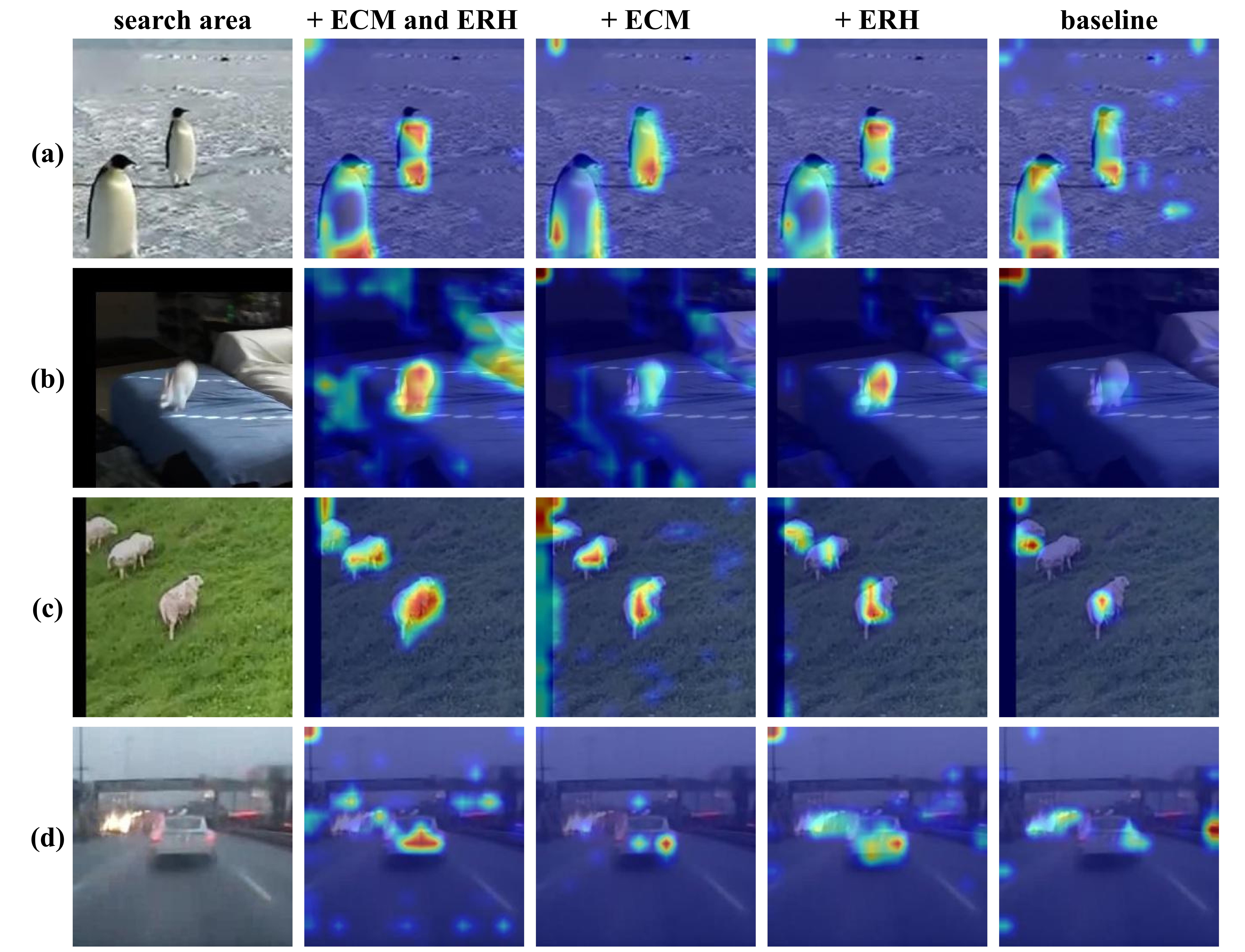}
	\caption{Visualization of heat maps of different LightFC models on LaSOT based on Grad-CAM. (a) bird-2 (frame 77), (b) rabbit-10 (frame 294), (c) sheep-9 (frame 61), (d) car-2(frame 159).}
	\label{fig:fig9}
\end{figure}

Fig. \ref{fig:fig9} shows heat maps of the baseline of LightFC, LightFC with ECM only, LightFC with ERH only, and LightFC. Heat maps visualize the output features of the feature fusion module. Both ECM and ERH increase the focus on key features in the model, respectively. Besides, they also jointly improve model feature representation. Overall, ECM and ERH effectively improves the feature representation of LightFC, proving the effectiveness of this work.

\subsubsection{Tracking Results}
\begin{figure}
	\centering
        \includegraphics[width=10cm]{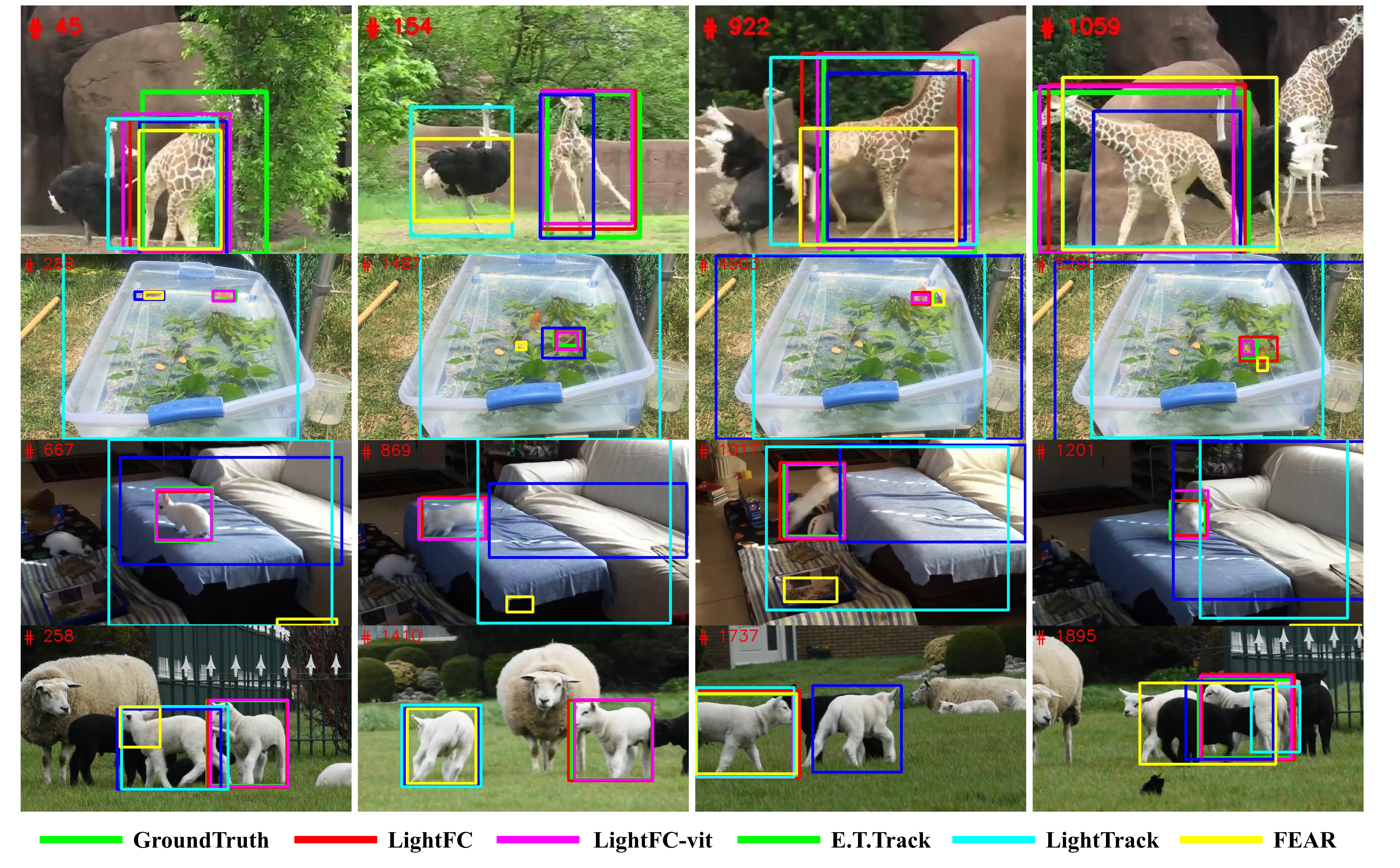}
	\caption{Visualized comparisons of LightFC and LightFC-vit with other lightweight trackers on four sequences of LaSOT. From top to bottom: giraffe-2, goldfish-7, rabbit-10, sheep-3.}
	\label{fig:fig10}
\end{figure}

Fig. \ref{fig:fig10} shows some tracking results of LightFC and LightFC-vit and other previous state-of-the-art lightweight trackers on four sequences of LaSOT [3]. In the giraffe-2 sequence, LightFC and LightFC-vit achieve more accurate bounding box prediction compared to other lightweight trackers. In the goldfish-7 sequence, when the target is partially occluded, LightFC and LightFC vit can still successfully locate targets, while the other lightweight trackers fail to locate the target. LightFC and LighFC-vit outperform other lightweight trackers in the rabbit-10 sequence by overcoming challenges such as deformation, fast motion, and motion blur. In the sheep-3 sequences, LightFC and LightFC-vit show better adaptability to challenges of similar objects and partial occlusion.

\section{Conclusion}
This paper proposes LightFC, a lightweight fully convolutional tracker. It abandons the heavy and expensive attention network structure to reduce parameters and Flops, and focuses on enhancing the model’s nonlinearity to improve the convolutional tracking pipeline’s performance. Specifically, we propose an efficient cross-correlation module (ECM) and an efficient reparameterization head (ERH). Comprehensive experiments show that: (1) It is reasonable to design the architecture of the ECM based on the analysis of commonalities in pixel correlation and self-attention relationship modeling. The introduction of SCF unit and IAB can effectively enhance the feature representation and feature nonlinearity. Furthermore, incorporating reasonable skip connections and reusing search area features can further improve model performance. (2) The use of reparameterization techniques and the addition of a channel attention module effectively improve the feature representation bottleneck in standard center-heads. (3) Compared with current lightweight trackers, LightFC and LightFC-vit achieve state-of-the-art performance and the optimal balance between performance, parameters, Flops and FPS.

\section{Acknowledgments}
This research was funded by the National Natural Science Foundation of China, grant number 52371350, by the Natural Science Foundation of Hainan Province, grant number 2021JJLH0002, by the Foundation of National Key Laboratory, grant number JCKYS2022SXJQR-06 and 2021JCJQ-SYSJJ-LB06912.

\bibliographystyle{unsrt}
\bibliography{references}

\end{document}